\documentclass[10pt,twocolumn,letterpaper]{article}

\usepackage[pagenumbers]{cvpr} %

\input{preamble}

\usepackage[font=small]{caption}   %
\usepackage{graphicx}  %
\usepackage{array}
 \usepackage{multirow} 
\usepackage{calc,colortbl}
 \usepackage{tabularx, makecell, booktabs, adjustbox, pifont}
\newcolumntype{x}[1]{>{\centering\arraybackslash}p{#1}}
\newcolumntype{y}[1]{>{\raggedright\arraybackslash}p{#1}}
\newcolumntype{z}[1]{>{\raggedleft\arraybackslash}p{#1}}

\definecolor{cvprblue}{rgb}{0.21,0.49,0.74}
\usepackage[pagebackref,breaklinks,colorlinks,allcolors=cvprblue]{hyperref}

\def\paperID{1134} %
\def\confName{CVPR}
\def\confYear{2026}

\newcommand{\nameCOLOR}[1]{\textcolor{black}{#1}}
\newcommand{\nameMethod}{\nameCOLOR{\mbox{SpaceTimePilot}}\xspace}
\newcommand{\nameDataset}{\nameCOLOR{\mbox{Cam$\times$Time}}\xspace}
\newcommand{\recammaster}{\nameCOLOR{\mbox{ReCamMaster}}\xspace}
\newcommand{\syncammaster}{\nameCOLOR{\mbox{SynCamMaster}}\xspace}
\newcommand{\duster}{\nameCOLOR{\mbox{DUSt3R}}\xspace}
\newcommand{\dit}{\nameCOLOR{\mbox{DiT}}\xspace}
\newcommand{\fourdim}{\nameCOLOR{\mbox{4DiM}}\xspace}
\newcommand{\catfourd}{\nameCOLOR{\mbox{CAT4D}}\xspace}
\newcommand{\suppmat}{\nameCOLOR{\mbox{Supp.~Mat.}}\xspace}

\newcommand{\src}{\text{src}}
\newcommand{\trg}{\text{trg}}
\newcommand{\prv}{\text{prv}}
\newcommand{\sourcevideo}{V_{\src}}
\newcommand{\targetvideo}{V_{\trg}}
\newcommand{\camera}{\mathbf{c}}
\newcommand{\animatime}{\mathbf{t}}
\newcommand{\numf}{F}
\newcommand{\numchannel}{C}
\newcommand{\width}{W}
\newcommand{\height}{H}
\newcommand{\videospace}{\mathbb{R}^{\numf \times \numchannel \times \height \times \width}}
\newcommand{\numflt}{F'}

\newcommand{\camencoder}{\mathcal{E}_\text{cam}}
\newcommand{\animaencoder}{\mathcal{E}_\text{ani}}

\newcommand{\todo}[1]{{\color{black}#1}}

\definecolor{darkgreen}{rgb}{0, 0.5, 0}
\definecolor{darkred}{rgb}{0.6, 0.1, 0.05}
\newcommand{\cmark}{\ding{52}}
\newcommand{\greencmark}{\color{darkgreen}{\cmark}}
\newcommand{\xmark}{\ding{56}}
\newcommand{\redxmark}{\color{darkred}{\xmark}}

\usepackage{etoc}

\newcommand{\logo}[1][2.2em]{%
  \raisebox{-0.4\height}{\includegraphics[height=#1]{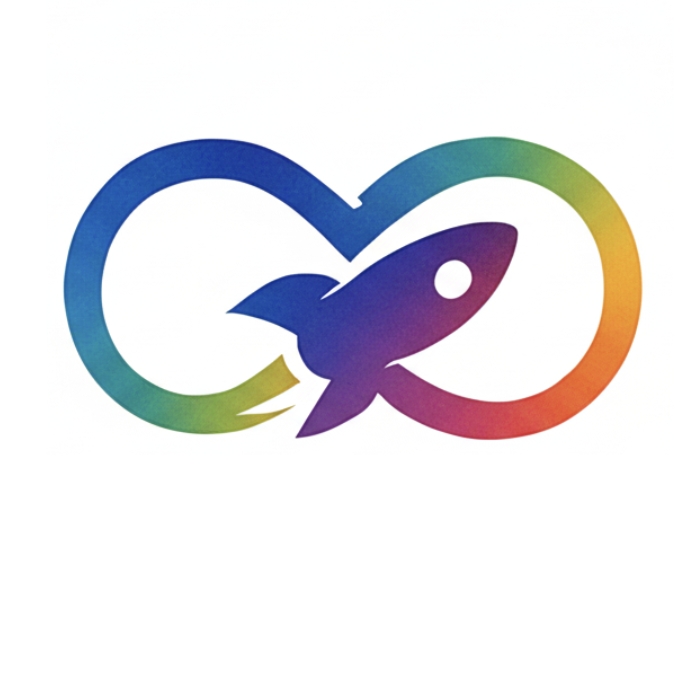}}%
}

\title{\logo \space \nameMethod: Generative Rendering of Dynamic Scenes Across \\  Space and Time

}

\author{
Zhening Huang$^{1,2}$ \quad 
Hyeonho Jeong$^2$ \quad 
Xuelin Chen$^2$ \quad 
Yulia Gryaditskaya$^2$ \\
Tuanfeng Y. Wang$^2$ \quad
Joan Lasenby$^1$ \quad 
Chun-Hao Huang$^2$ \\[1.5ex]
$^1$University of Cambridge \qquad $^2$Adobe Research \\[1.5ex]
\texttt{\url{https://zheninghuang.github.io/Space-Time-Pilot/}} 
}

\begin{document}

\twocolumn[{%
\renewcommand\twocolumn[1][]{#1}%
\vspace{-12mm}
\begin{center}
\maketitle
\vspace{-10mm}
\captionsetup{type=figure}
\makebox[\linewidth][c]{\includegraphics[width=1.1\linewidth]{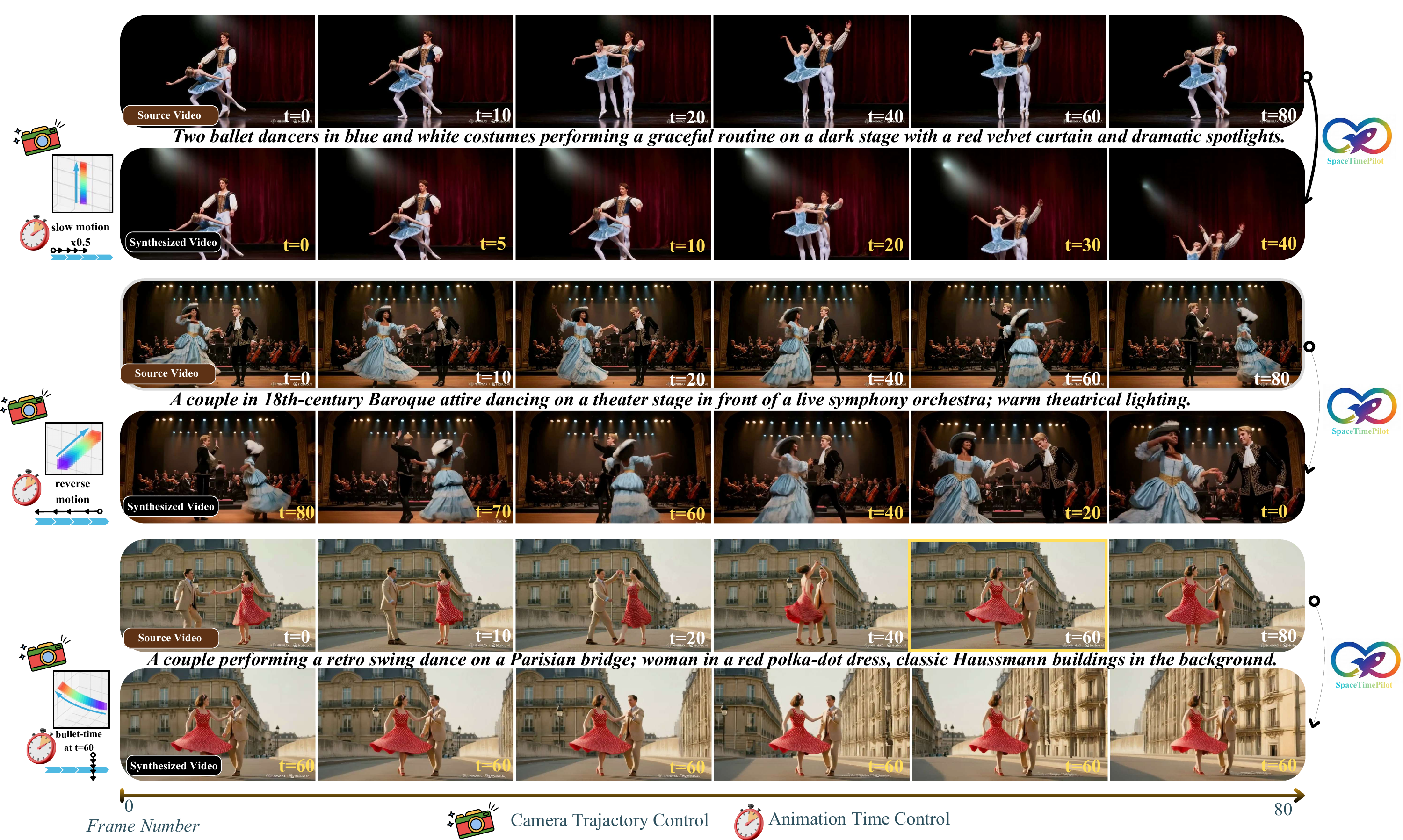}}
\captionof{figure}{
\textbf{SpaceTimePilot} enables unified control over both camera and time within a single diffusion model, producing continuous and coherent videos along arbitrary space–time trajectories. Given a source video (odd rows), our model synthesizes new videos (even rows) with retimed motion sequences, including slow motion, reverse motion, and bullet time, while precisely controlling camera movement according to a given camera trajectory.
}
\label{fig:teaser}
\end{center}
}]

\begin{abstract}

We present \textbf{\nameMethod}, a video diffusion model that disentangles space and time for controllable generative rendering. Given a monocular video, \textbf{\nameMethod} can independently alter the camera viewpoint and the motion sequence within the generative process, re-rendering the scene for continuous and arbitrary exploration across space and time.
To achieve this, we introduce an effective animation time-embedding mechanism in the diffusion process, allowing explicit control of the output video’s motion sequence with respect to that of the source video. As no datasets provide paired videos of the same dynamic scene with continuous temporal variations, we propose a simple yet effective \textbf{temporal-warping training scheme} that repurposes existing multi-view datasets to mimic temporal differences. This strategy effectively supervises the model to learn temporal control and achieve robust space–time disentanglement. 
To further enhance the precision of dual control, we introduce two additional components: an improved camera-conditioning mechanism that allows altering the camera from the first frame, and \nameDataset, the first synthetic Space and Time full-coverage rendering dataset that provides fully free space–time video trajectories within a scene. Joint training on the temporal-warping scheme and the \nameDataset dataset yields more precise temporal control.
We evaluate \textbf{\nameMethod} on both real-world and synthetic data, demonstrating clear space–time disentanglement and strong results compared to prior work.
\end{abstract}
    
\section{Introduction}
\label{sec:intro}

Videos are 2D projections of an evolving 3D world, where the underlying generative factors consist of spatial variation (camera viewpoint) and temporal evolution (dynamic scene motion). Learning to understand and disentangle these factors from observed videos is fundamental for tasks such as scene understanding, 4D reconstruction, video editing, and generative rendering, to name a few.  In this work, we approach this challenge from the perspective of generative rendering. Given a single observed video of a dynamic scene, our goal is to synthesize novel views (reframe/reangle) and/or at different moments in time (retime), while remaining faithful to the underlying scene dynamic. 

A common strategy is to first reconstruct dynamic 3D content from 2D observations, \ie, perform 4D reconstruction, and then re-render the scene. 
These methods model both spatial and temporal variations using representations such as NeRFs \cite{lin2024dynamic,mildenhall2021nerf} or Dynamic Gaussian Splatting \cite{kerbl20233d,wu20244d}, often aided by cues like geometry \cite{park2021hypernerf,park2021nerfies}, optical flow \cite{li2023dynibar,li2021nsff}, depth \cite{gao2021dynamic,yoon2020novel}, or long-term 2D tracks \cite{wang2024shapeofmotion,lei2024mosca}. However, even full 4D reconstructions typically show artifacts under novel viewpoints.
More recent work \cite{Wu2024, liang2024diffusion4d} uses multi-view video diffusion to generate sparse, time-conditioned views and refines them via Gaussian-splatting optimization, but rendering quality remains limited.
Advances in video diffusion models \cite{zhang2024show, wang2024lavie, blattmann2023stable, jeong2024track4gen, bar2024lumiere, videoworldsimulators2024, yang2024cogvideox, polyak2024moviegencastmedia, kong2024hunyuanvideo} further enable camera re-posing with more lightweight point cloud representations, reducing the need for heavy 4D reconstruction. While effective in preserving identity, their reliance on per-frame depth and reprojection limits robustness under large viewpoint changes. To mitigate this, newer approaches condition generation solely on camera parameters, achieving strong novel-view synthesis on both static \cite{jin2025lvsm} and dynamic scenes \cite{Bai2025,vanhoorick2024gcd,He2025}. 
Autoregressive models like Genie-3 \cite{parker-holder2025genie3} even enable interactive scene exploration from a single image, showing that diffusion models can encode implicit 4D priors. 
Nonetheless, despite progress in spatial viewpoint control, current methods still lack full 4D exploration, \ie, the ability to navigate scenes freely across both space and time.

\begin{figure}[t]
    \centering
    \includegraphics[width=\linewidth]{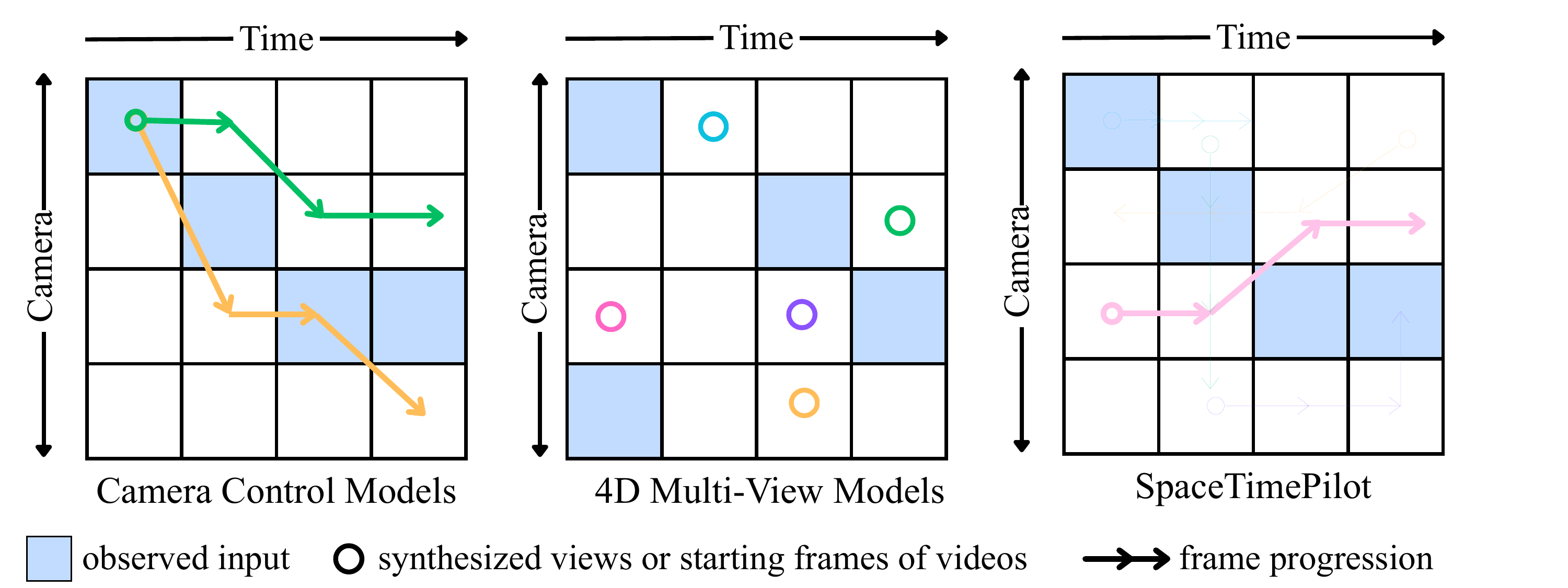}

\caption{\textbf{Space–time controllability across methods.}
Blue cells denote the input video/views, while arrows and dots indicate generated continuous videos or sparse frames. 
Camera-control V2V models~\cite{Bai2025, vanhoorick2024gcd} modify only the camera trajectory while keeping time strictly monotonic. 
4D multi-view models~\cite{Wu2024, liang2024diffusion4d} synthesize discrete sparse views conditioned on space and time, but do not generate continuous video sequences. 
\textit{SpaceTimePilot} enables free movement along both the camera and time axes with full control over direction and speed, supporting bullet-time, slow-motion, reverse playback, and mixed space–time trajectories.}

    \label{fig:high_level}
    \vspace{-3mm}
\end{figure}

In this work, we introduce \textit{\nameMethod}, the first video diffusion model that enables joint spatial and temporal control. \nameMethod introduces a new notion of ``animation time'' to capture the \textit{temporal status} of scene dynamics in the source video. As such, it naturally disentangles temporal control and camera control by expressing them as two independent signals. 
A high-level comparison between our approach and prior methods is illustrated in Fig.~\ref{fig:high_level}. Unlike previous methods, \nameMethod enables free navigation along both the camera and time axes.
Training such a model requires dynamic videos that exhibit multiple forms of temporal playback while simultaneously being captured under multiple camera motions, which is only feasible in a controlled studio setups. 
Although temporal diversity can be increased by combining multiple real datasets \eg \cite{zhou2018stereo,ling2024dl3dv}, as done in \cite{watson2025controllingspacetimediffusion,Wu2024}, this approach remains suboptimal, as the coverage of temporal variation is still insufficient to learn the underlying meaning of temporal control.
Existing synthetic datasets \cite{Bai2025, bai2024syncammaster} also do not exhibit such properties.

To address this limitation, we introduce a simple yet effective \textit{temporal-warping training scheme} that augments existing multi-view video datasets \cite{Bai2025, bai2024syncammaster} to simulate diverse conditioning types while preserving continuous video structure. By warping input sequences in time, the model is exposed to varied temporal behaviors without requiring additional data collection. This simple yet crucial strategy allows the model to learn temporal control signals, enabling it to directly exhibit space–time disentanglement effects during generation.
We further ablate various temporal-conditioning schemes and introduce a convolution-based temporal-control mechanism that enables finer-grained manipulation of temporal behavior and supports effects such as bullet-time at any timestep within the video.
While temporal warping increases temporal diversity, it can still entangle camera and scene dynamics -- for example, temporal manipulation may inadvertently affect camera behavior. 
To further strengthen disentanglement, we introduce a new dataset that spans the full grid of camera–time combinations along a trajectory. Our synthetic \textit{\nameDataset} dataset contains 180k videos rendered from 500 animations across 100 scenes and three camera paths. Each path provides full-motion sequences for every camera pose, yielding dense multi-view and full-temporal coverage. This rich supervision enables effective disentanglement of spatial and temporal control.

Experimental results show that \nameMethod successfully disentangles space and time in generative rendering from single videos, outperforming adapted state-of-the-art baselines by a significant margin. 
Our main contributions are summarized as follows:
\begin{itemize}
\item	We introduce \nameMethod, the first video diffusion model that disentangles spatial and temporal factors to enable continuous and controllable novel view synthesis as well as temporal control from a single video.
\item  We propose the \textit{temporal-warping strategy} that repurposes multi-view datasets to simulate diverse temporal variations. By training on these warped sequences, the model effectively learns temporal control without the need for explicitly constructed video pairs captured under different temporal settings.
\item	We propose a more precise camera–time conditioning mechanism, illustrating how viewpoint and temporal embeddings can be jointly integrated into diffusion models to achieve fine-grained spatiotemporal control.
\item	We construct the \nameDataset Dataset, providing dense spatiotemporal sampling of dynamic scenes across camera trajectories and motion sequences. This dataset supplies the necessary supervision for learning disentangled 4D representations and supports precise camera–time control in generative rendering.
\end{itemize}

\section{Related work}

We aim to re-render a video from new viewpoints with temporal control, a task closely related to Novel View Synthesis (NVS) from monocular video inputs.

\myparagraph{Video-based NVS.}
Prior video-based NVS methods can be broadly characterized along two axes: (i) whether they target static or dynamic scenes, and (ii) whether they incorporate explicit 3D geometry in the generation pipeline.

For static scenes, geometry-based methods reconstruct scene geometry from the input frames and use diffusion models to complete or hallucinate regions that are unseen under new viewpoints \cite{yu2024viewcrafter,Ren2025,Jeong2025,wu2025video}.
Although these approaches achieve high rendering quality, they rely on heavy 3D preprocessing.
Geometry-free approaches \cite{Bai2025,Zhou2025,vanhoorick2024gcd} bypass explicit geometry and directly condition the diffusion process on observed views and camera poses to synthesize new viewpoints.

For dynamic scenes, inpainting-based methods such as TrajectoryCrafter \cite{YU2025}, ReCapture \cite{Zhang2024}, and Reangle \cite{Jeong2025} also adopt warp-and-inpaint pipelines, while GEN3C \cite{Ren2025} extends this with an evolving 3D cache and EPiC \cite{zun2025epic} improves efficiency via a lightweight ControlNet framework.
Geometry-free dynamic models \cite{Bai2025,bai2024syncammaster,vanhoorick2024gcd, wang20254realv2, Wang_2025_CVPR} instead learn camera-conditioned generation from multi-view or 4D datasets (\eg, Kubric-4D \cite{greff2021kubric}), enabling smoother and more stable NVS with minimal 3D inductive bias.
Proprietary systems like Genie 3 \cite{parker-holder2025genie3} further demonstrate real-time, continuous camera control in dynamic scenes, underscoring the potential of video diffusion models for interactive viewpoint manipulation.

\myparagraph{Disentangling Space and Time.}
Despite great progress in camera controllability (space), 
the methods discussed above do not address temporal control (time).
Meanwhile, disentangling spatial and temporal factors has become a central focus in 4D scene generation, recently advanced through diffusion-based models.
\fourdim \cite{watson2025controllingspacetimediffusion} introduces a Masked FiLM mechanism that defaults to identity transformations when conditioning signals (e.g., camera pose or time) are absent, enabling unified representations across both static and dynamic data through multi-modal supervision. Similarly, \catfourd \cite{Wu2024} leverages multi-view images to conduct 4D dynamic reconstruction to achieve space–time disentanglement but remains constrained by its reliance on explicit 4D reconstruction pipelines, which limits scalability and controllability. 
In contrast, our approach builds upon text-to-video diffusion models and introduces a new temporal embeddings module and refined camera conditioning to achieve fully controllable 4D generative reconstruction.

\section{Method}
\label{sec:method}

We introduce \nameMethod, a method that takes a source video $\sourcevideo \in \videospace$ as input and synthesizes a target video $\targetvideo \in \videospace$, following an input camera trajectory $\camera_{\trg}\in \mathbb{R}^{\numf \times 3 \times 4}$ and temporal control signal $\animatime_{\trg} \in \mathbb{R}^{\numf}$.
Here, $\numf$ denotes the number of frames, $\numchannel$ the number of color channels, 
and $\height$ and $\width$ are the frame height and width, respectively. 
Each $\camera_{\trg}^{f} \in \mathbb{R}^{3 \times 4}$ represents the camera extrinsic 
parameters (rotation and translation) at frame $f$, with respect to the 1\textsuperscript{st} frame of  $\sourcevideo$.
The target video $\targetvideo$ preserves the scene’s underlying dynamics, geometry, and appearance in $\sourcevideo$, while adhering to the camera motion and temporal progression specified by $\camera_{\trg}$ and $\animatime_{\trg}$.
A key feature of our method is the disentanglement of spatial and temporal factors in the generative process, enabling effects such as bullet-time and retimed playback from novel viewpoints (see \cref{fig:teaser}).

\subsection{Preliminaries}
Our framework builds upon recent advances in large-scale text-to-video diffusion models and camera-conditioned video generation. 
We adopt a latent video diffusion backbone similar to modern text-to-video foundation models \cite{wan2025}, consisting of a 3D Variational Auto-Encoder (VAE) for latent compression and a Transformer-based denoising model (DiT) operating over multi-modal tokens. 

Additionally, our design draws inspiration from \recammaster~\cite{Bai2025}, which introduces explicit camera conditioning for video synthesis. 
Given an input camera trajectory $\camera \in \mathbb{R}^{\numf \times 3 \times 4}$, 
spatial conditioning is achieved by first projecting the camera sequence to the space of video tokens and adding it to the features:
\begin{equation}
x' = x + \camencoder\left(\camera\right),
\label{eq:camera-condition}
\end{equation}
where $x$ is the output of the patchifying module and $x'$ is the input to self-attention layers. 
The camera encoder $\camencoder$ maps each flattened $3 \times 4$ camera matrix (12-dimensional) into the target feature space, while also transforming the temporal dimension from $\numf$ to $\numflt$.

\subsection{Disentangling Space and Time}
We achieve spatial and temporal disentanglement through a two-fold approach: a dedicated time representation and specialized datasets.

\subsubsection{Time representation}
\label{sec:time_embedding}
Recent video diffusion models include position embeddings for latent frame index $f'$, such as RoPE($f'$). 
However, we found using RoPE($f'$) for temporal control to be ineffective, as it interferes with camera signals: RoPE($f'$) often constrains both temporal and camera motion simultaneously.
To address space and time disentanglement, we introduce a dedicated time control parameter $\animatime \in \mathbb{R}^F$.
By manipulating $\animatime_\trg$, we can control the temporal progression of the synthesized video $\targetvideo$. For example, setting $\animatime_\trg$ to a constant locks $\targetvideo$ to a specific timestamp in $\sourcevideo$, while reversing the frame indices produces a playback of $\sourcevideo$ in reverse.
\begin{figure}[t]
    \centering
    \includegraphics[width=\linewidth]{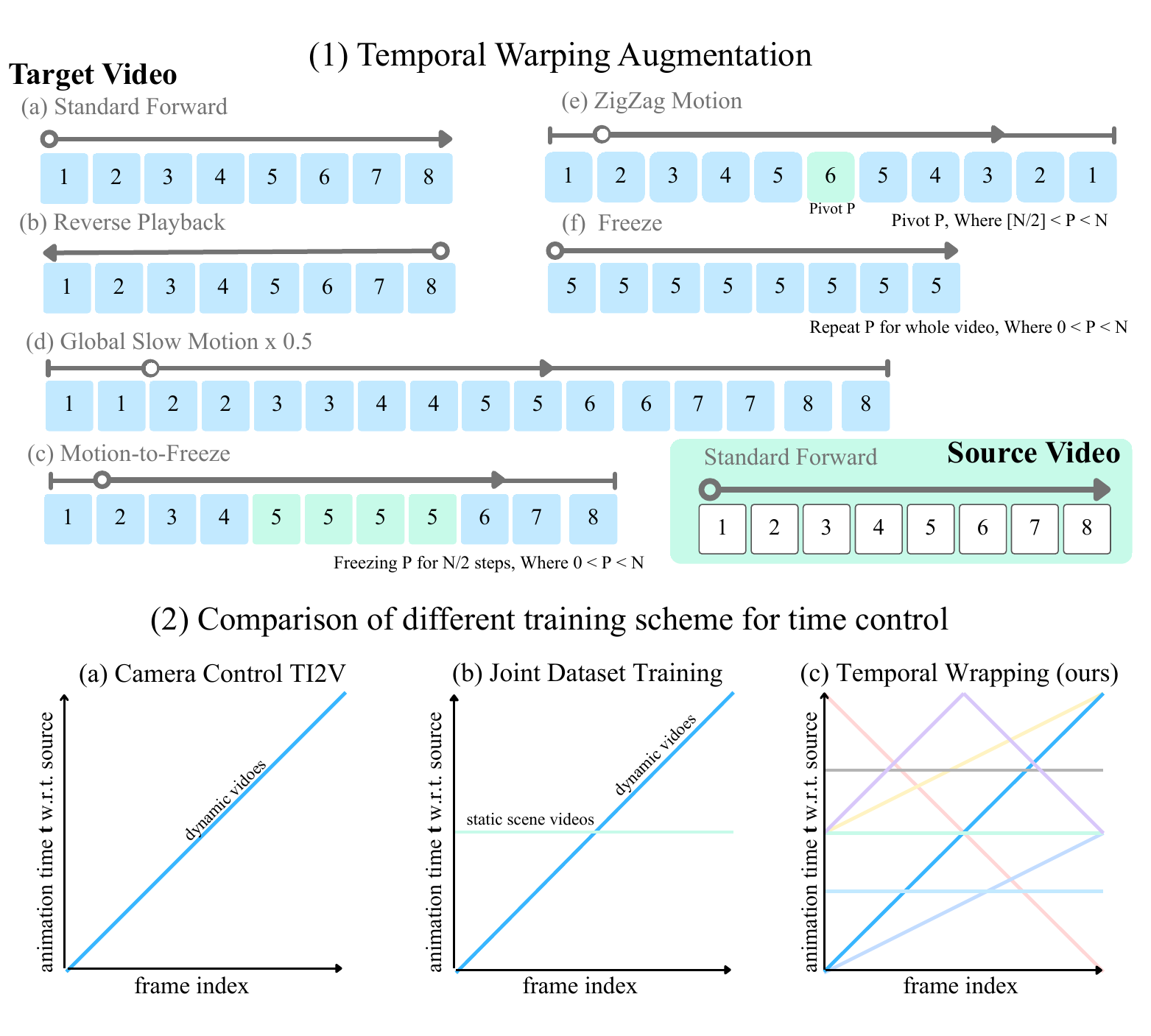}
    \vspace{-0.5cm}
    \caption{
        \textbf{Temporal Wrapping for Spatiotemporal Disentanglement.}
        (Top) For multi-view dynamic scene datasets \cite{Bai2025}, a set of temporal warping operations (e.g.\, reverse playback, zigzag motion, slow motion, and freeze) are applied to the target video, with the source video kept as the standard forward reference, providing explicit supervision for temporal control  .
        (Bottom) Compared with existing camera-control \cite{vanhoorick2024gcd, Bai2025} and joint-dataset training strategies \cite{Wu2024, watson2025controllingspacetimediffusion}, which rely on monotonic time progression and static-scene videos to demonstrate temporal differences, Temporal Wrapping provide much more diverse and explicit signals of temporal variation, leading to disentanglement of space and time.
    }
    \label{fig:temporal_wrapping}
    \vspace{-0.3cm}
\end{figure}

(Top) For multi-view dynamic scene datasets, a set of temporal warping operations, including reverse, playback, zigzag motion, slow motion, and freeze are apppplied with teh source video as standford. This gives explicit supervision for temporal control, without constructing additional temporally varied training data.

(Bottom) Existing camera-control and joint dataset training rely on monotonic time progression and static scene videos, making it difficult for models to understand temporal variation. The introduced temporal mappings from multi-view video data, which provide diverse and clear signal on tempral variation, and directly lead to disentanglement of space and time.

\myparagraph{Time Embedding.}
To inject temporal control into the diffusion model, we analyze several approaches. 
First, we can encode time similar to a frame index using RoPE embedding. 
However, we find it less suitable for time control (\todo{visual evaluations are provided in \suppmat}).
Instead, we adopt sinusoidal time embeddings applied at the latent frame $f'$ level, which provide a stable and continuous representation of each frame’s temporal position and offer a favorable trade-off between precision and stability. 
We further observe that each latent frame corresponds to a continuous temporal chunk, and propose using embeddings of original frame indices $f$ to support finer granularity of time control. 
To accomplish this, we introduce a time encoding approach $\animaencoder(\animatime)$, where $\animatime \in \mathbb{R}^F$. 
We first compute the sinusoidal time embeddings to represent the temporal sequence, $\mathbf{e}_\src = \mathrm{SinPE}\!\left(\animatime_\src\right)$, $\mathbf{e}_\trg = \mathrm{SinPE}\!\left(\animatime_\trg\right)$, where  $\animatime_\src, \animatime_\trg \in \mathbb{R}^F$.
Next, we apply two 1D convolution layers to progressively project these embeddings into the latent frame space, $\widetilde{\mathbf{e}} = \mathrm{Conv1D_2}(\mathrm{Conv1D_1}(\mathbf{e}))$.
Finally, we add these time features to the camera features and video tokens embeddings, updating \cref{eq:camera-condition} as follows:
\begin{equation}
    x' = x + \camencoder\left(\camera\right) + \animaencoder\left(\animatime\right).
    \label{eq:cam-and-ani-condition}
\end{equation}

In \cref{sec:time_embedding_ablation}, we compare our approach with alternative conditioning strategies, such as using sinusoidal embeddings where $\animatime_\src,\animatime_\trg$ are directly defined in $\mathbb{R}^{F'}$, and employing an MLP instead of a 1D convolution for compression. We demonstrate both qualitatively and quantitatively the advantages of our proposed method.

\subsubsection{Datasets}
\label{sec:datasets}
To enable temporal manipulation in our approach, we require paired training data that includes examples of time remapping. 
Achieving spatial-temporal disentanglement further requires data containing examples of both camera and temporal controls. 
To the best of our knowledge, no publicly available datasets satisfy these requirements.
Only a few prior works, such as \fourdim \cite{watson2025controllingspacetimediffusion} and \catfourd \cite{Wu2024}, have attempted to address spatial-temporal disentanglement. 
A common strategy is to jointly train on static-scene datasets and multi-view video datasets \cite{zhou2018stereo,ling2024dl3dv}. 
The limited control variability in these datasets leads to confusion between temporal evolution and spatial movement, resulting in entangled or unstable behaviors \cite{watson2025controllingspacetimediffusion,Wu2024}. 
We address this limitation by augmenting existing multi-view video data with temporal warping and by proposing a new synthetic dataset.

\myparagraph{Temporal Warping Augmentation.}
We introduce simple augmentations that add controllable temporal variations to multi-view video datasets.
During training, given a source video $\sourcevideo = \{I_\src^f\}_{f=1}^{F}$ and a target video $\targetvideo = \{I_\trg^f\}_{f=1}^{F}$, we apply a temporal warping function $\tau: [1, \numf] \rightarrow [1, \numf]$ to the target sequence, producing a warped video $V_\trg' = \{I_\trg^{\tau(f)}\}_{f=1}^{F}$. 
The source animation timestamps are uniformly sampled, $\animatime_\src = 1:F$. 
Warped timestamps, $\animatime_\trg = \tau(\animatime_\src)$, introduce non-linear temporal effects (see \cref{fig:temporal_wrapping} top b–e): (i) reversal, (ii) acceleration, (iii) freezing, (iv) segmental slow motion, and (v) zigzag motion, in which the animation repeatedly reverses direction.
After these augmentations, the paired video sequences $(V_\src, V_\trg')$ differ in both camera trajectories and temporal dynamics, providing the model with a clear signal for learning disentangled spatiotemporal representations.

\begin{table}[!t]
  \centering
  \definecolor{Gray}{gray}{0.9}
  \caption{
\textbf{Comparison of existing multi-view datasets for camera and temporal control against Cam$\times$Time.} Cam$\times$Time provides full-grid rendering (Figure~\ref{fig:dataset_blender}), enabling target videos to sample arbitrary temporal variations over the full range from 0 to 120.
  }
  \vspace{-0.2cm}
  \label{tab:dataset-comparison}
  \begin{adjustbox}{max width=\columnwidth}
  \begin{tabular}{l|c|c|c|c}
    \toprule
     Dataset & Dynamic scenes & Src. Time: $t_{\text{src}}$ & Tgt. Time: $t_{\text{trg}}$ & Camera \\
    \midrule
    \midrule
    RE10k \cite{zhou2018stereo} & \redxmark & 1 & 1 & Moving \\
    DL3DV10k \cite{ling2024dl3dv} & \redxmark & 1 & 1 & Moving \\
    MannequinChallenge \cite{rockwell2025dynpose} & \redxmark & 1 & 1 & Moving \\
    Kubric-4D \cite{vanhoorick2024gcd} & \greencmark & 1:60 & 1:60 & Moving \\
    ReCamMaster \cite{Bai2025} & \greencmark & 1:80 & 1:80 & Moving \\
    SynCamMaster \cite{bai2024syncammaster} & \greencmark & 1:80 & 1:80 & Fixed \\
    \midrule
        \rowcolor{Gray}
    \textbf{Cam$\times$Time (ours)} & \greencmark & 1:120 & $\{1, 2, \dots, 120\}^{120}$ & Moving \\
    \bottomrule
  \end{tabular}
  \end{adjustbox}
  \vspace{-3.5mm}
\end{table}

\begin{figure}[t]
    \centering
    \includegraphics[width=\columnwidth,trim={0 0 0 3mm},clip]{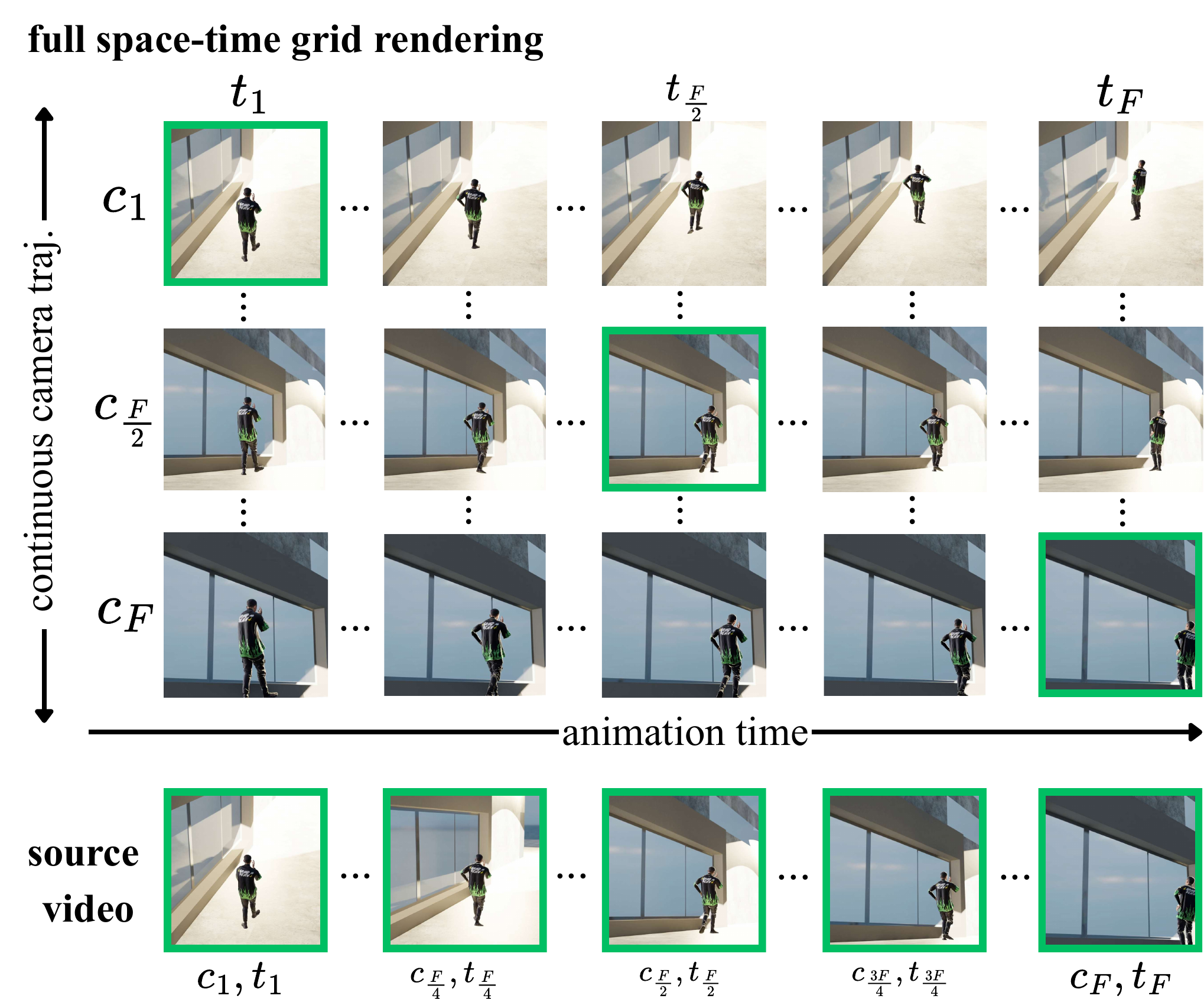}
    \caption{
    \textbf{\textit{\nameDataset} dataset visualization}. (Top) A space-time grid defined by a camera trajectory $\camera=[c_1,...,c_F]$ and animation status $\animatime=[t_1,...,t_F]$. \nameDataset renders images for all $(c,t)$ pairs, covering the full grid for learning disentangled spatial and temporal control. Any two sampled  sequences of $\numf$ frames from the grid can form a source-target pair. (Bottom) One typical choice of source videos is taking the diagonal cells in     green. }
    \label{fig:dataset_blender}
    \vspace{-0.3cm}
\end{figure}

\myparagraph{Synthetic \nameDataset Dataset for Precise Spatiotemporal Control.}
While our temporal warping augmentations encourage strong disentanglement between spatial and temporal factors, achieving fine-grained and continuous control --- that is, smooth and precise adjustment of temporal dynamics --- benefits from a dataset that systematically covers both dimensions.
To this end, we construct \textit{\nameDataset}, a new synthetic spatiotemporal dataset rendered in Blender. 
Given a camera trajectory and an animated subject, \nameDataset exhaustively samples the camera–time grid, capturing each dynamic scene across diverse combinations of camera viewpoints and temporal states $\left(\camera, \animatime\right)$, as illustrated in \cref{fig:dataset_blender}.
The source video is obtained by sampling the diagonal frames of the dense grid (\cref{fig:dataset_blender} (bottom)), while the target videos are obtained by more free-form sampling of continuous sequences. 
We compare \nameDataset against existing datasets in \cref{tab:dataset-comparison}. 
While \cite{zhou2018stereo,ling2024dl3dv,rockwell2025dynpose} are real videos with complex camera path annotations, they either do not provide time-synchronized video pairs \cite{rockwell2025dynpose} or only provide pairs of static scenes \cite{ling2024dl3dv,zhou2018stereo}.
Synthetic multi-view video datasets \cite{Bai2025,bai2024syncammaster,vanhoorick2024gcd} provide pairs of dynamic videos but do not allow training for time control. 
In contrast, \nameDataset enables fine-grained manipulation of both camera motion and temporal dynamics, enabling bullet-time effects, motion stabilization, and flexible combinations of the controls.
We designate part of \nameDataset as a test set, aiming for it to serve as a benchmark for controllable video generation.
We will release it to support future research on fine-grained spatiotemporal modeling.

\begin{table*}[t]
    \centering
    \caption{
        Quantitative comparison across temporal controls 
        (\textit{Direction (forward, backward motion)}, \textit{Speed (slow modes)}, \textit{Bullet Time}).
        We report PSNR$\uparrow$, SSIM$\uparrow$, and LPIPS$\downarrow$.
        Best results are in \textbf{bold}. SpaceTimeMethod showcase best performance for temporal control overall.
    }
    \vspace{-0.2cm}
    \label{tab:temporal_eval}
    \small
    \setlength\tabcolsep{3.6pt}  %
    \begin{tabular}{l|ccc|c|ccc|c|ccc|c}
        \toprule
        \multirow{2}{*}{\textbf{Method}} &
        \multicolumn{4}{c|}{\textbf{PSNR$\uparrow$}} &
        \multicolumn{4}{c|}{\textbf{SSIM$\uparrow$}} &
        \multicolumn{4}{c}{\textbf{LPIPS$\downarrow$}} \\
        \cmidrule(lr){2-5} \cmidrule(lr){6-9} \cmidrule(lr){10-13}
         & Dir. & Speed & Bullet & Avg
         & Dir. & Speed & Bullet & Avg
         & Dir. & Speed & Bullet & Avg \\
        \midrule

        ReCamM+preshuffled$^\dagger$
        & 17.13 & 14.84 & 14.61 & 15.52
        & 0.6623 & 0.6050 & 0.5965 & 0.6213
        & 0.3930 & 0.4793 & 0.4863 & 0.4529 \\

        ReCamM+jointdata
        & 18.32 & 17.57 & 17.69 & 17.86
        & 0.7322 & 0.7220 & 0.7209 & 0.7250
        & 0.2972 & 0.3158 & 0.3089 & 0.3073 \\

        \textbf{\nameMethod (Ours)}
        & \textbf{21.75} & \textbf{20.87} & \textbf{20.85} & \textbf{21.16}
        & \textbf{0.7725} & \textbf{0.7645} & \textbf{0.7653} & \textbf{0.7674}
        & \textbf{0.1697} & \textbf{0.1917} & \textbf{0.1677} & \textbf{0.1764} \\
        \bottomrule
    \end{tabular}

    {\footnotesize
    $^\dagger$\,Uses simple frame-rearrangement operators 
    (reversal, repetition, freezing) applied prior to inference to emulate 
    temporal manipulation.}
    \vspace{-0.3cm}
\end{table*}

\subsection{Precise Camera Conditioning}
\label{sec:src_cam}
We aim for full camera trajectory control in the target video.
In contrast, the previous novel-view synthesis approach \cite{Bai2025} assumes that the first frame is identical in source and target videos and that the target camera trajectory is defined relative to it. 
This stems from the two limitations. 
First, the existing approach ignores the source video trajectory, yielding suboptimal source features computed using the target trajectory for consistency:
\[
x_\src' = x_\src + \camencoder\left(\camera_\trg\right), \quad x_\trg' = x_\trg+ \camencoder\left(\camera_\trg\right).
\]
Second, it is trained on datasets where the first frame is always identical across the source and target videos. 
This latter limitation is addressed in our training datasets design.

To overcome the former, we devise a \textit{source-aware camera conditioning}. 
We estimate camera poses for both the source and target videos using a pretrained pose estimator, and inject them jointly into the diffusion model to provide explicit geometric context. Eq.~\ref{eq:cam-and-ani-condition} is therefore extended into:
\begin{align}
&x_\src' = x_\src + \camencoder\left(\camera_\src\right) + \animaencoder\left(\animatime_\src\right), \label{eq-srccam} \\
&x_\trg' = x_\trg + \camencoder\left(\camera_\trg\right) + \animaencoder\left(\animatime_\trg\right), \nonumber \\
&x' = [x_\trg', x_\src']_{\text{frame-dim}}, \nonumber
\end{align}
where $x'$ denotes the input of the \dit model, which is the concatenation of target and source tokens along the frame dimension. This formulation provides the model with both source and target camera context, enabling spatially consistent generation and precise control over camera trajectories.

\subsection{Support for Longer Video Segments}
\label{sec:LongerVideoSegements}
Finally, to showcase the full potential of our camera and temporal control, we adopt a simple autoregressive video generation strategy, generating each new segment $V_{\trg}$ conditioned on the previously generated segment $V_{\prv}$ and a source video $V_{\src}$ to produce longer videos. 

To enable this capability during inference, we need to extend our training scenario to support conditioning on two videos, where one serves as $V_{\src}$ and the other as $V_{\prv}$.
The source video $V_{\src}$ is taken directly from the multi-view datasets or from our synthetic dataset, as was described previously. 
$V_{\prv}$ is constructed in a similar way to $V_{\trg}$ --- either using temporal warping augmentations or by sampling from the dense space-time grid of our synthetic dataset. 
When temporal warping is applied, $V_{\prv}$ and $V_{\trg}$ may originate from the same or different multi-view sequences representing the same time interval. 
To maintain full flexibility of control, we do not enforce any other explicit correlations between $V_{\prv}$ and $V_{\trg}$, apart from specifying camera parameters relative to the selected source video frame.

Note that not constraining the source and target videos to share the same first frame (as discussed in \cref{sec:src_cam}) is crucial for achieving flexible camera control in longer sequences.
For instance, this design enables extended bullet-time effects: we can first generate a rotation around a selected point up to $45^\circ$ ($V_{\trg,1}$), and then continue from $45^\circ$ to $90^\circ$ ($V_{\trg,2}$).
Conditioning on two consecutive source segments allows the model to leverage information from newly generated viewpoints.
In the bullet-time example, conditioning on the previously generated video enables the model to incorporate information from all newly synthesized viewpoints, rather than relying solely on the viewpoint of the corresponding moment in the source video.

\section{Experiments}
\begin{figure*}[ht]
    \centering
    \vspace{-3mm}
    \includegraphics[width=\linewidth]{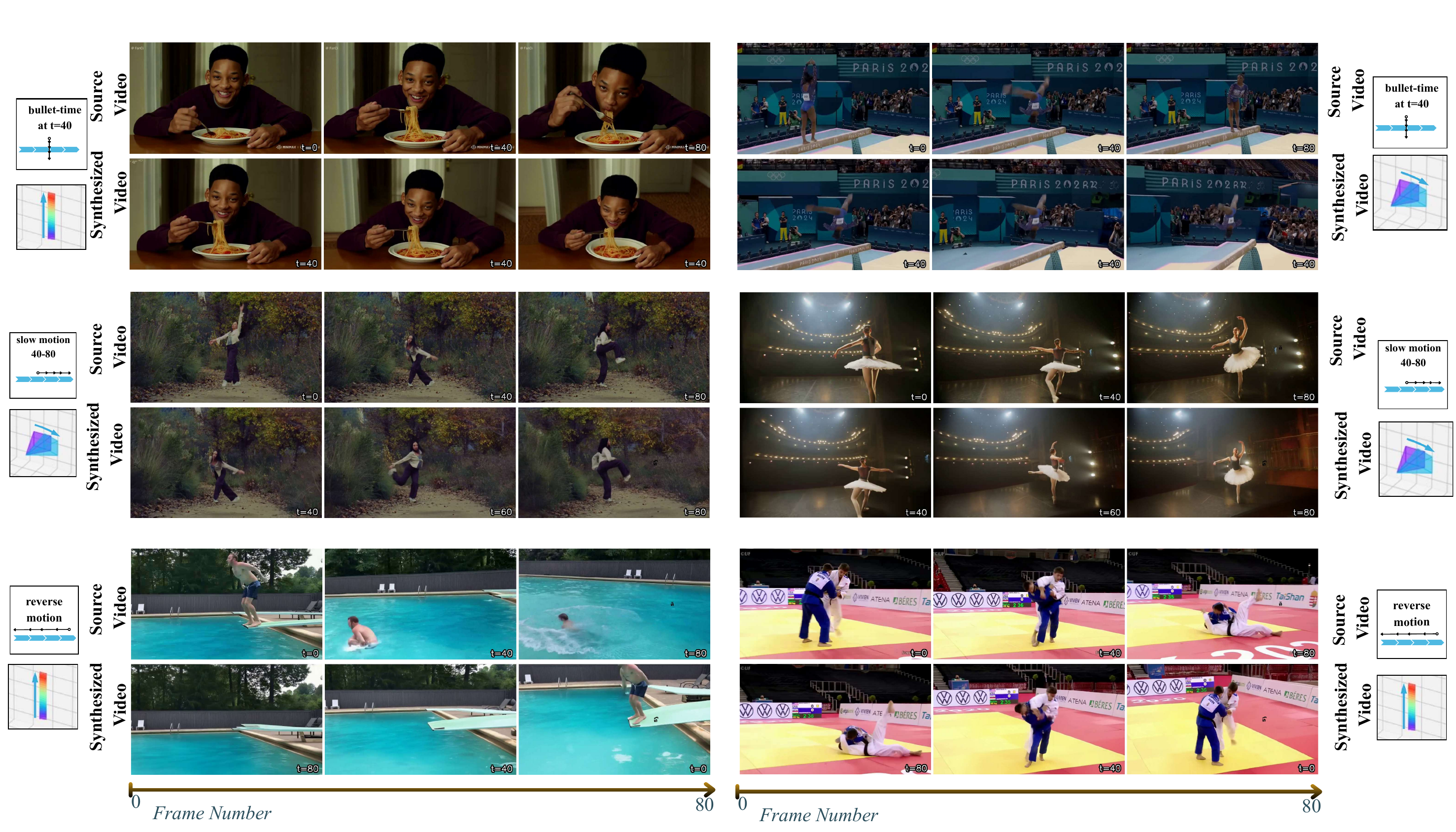}
    \caption{
    \textbf{Qualitative results of SpaceTimePilot.}
Our model enables fully disentangled control over camera motion and temporal dynamics.
Each row shows a different combination of camera trajectory (left icons) and temporal warping (right icons). SpaceTimePilot produces coherent videos under diverse controls, including normal playback, reverse playback, bullet-time, slow-motion, replay motion, and complex camera paths (pan, tilt, zoom, and vertical motion).
    }
    \label{fig:quali}
\end{figure*}

\myparagraph{Implementation details.}
We adopt the Wan-2.1 T2V-1.3B model \cite{wan2025}, which produces $\numflt$=21 latent frames and decodes them into $\numf$=81 RGB frames using a 3D-VAE.
The network is conditioned on camera and animation-time controls as defined in Eq.~\ref{eq-srccam}.  
Unless otherwise specified, \nameMethod is trained with \recammaster and \syncammaster datasets with the temporal warping augmentation described in Sec.~\ref{sec:datasets}, along with \nameDataset. Please refer to \suppmat for complete network architecture and additional training details.  
\subsection{Comparison with State-of-the-Art Baselines}
\subsubsection{Time-Control Evaluation.}
\label{sec:eval:time_control}
We first evaluate the retiming capability of our model. 
To factor out the error induced by camera control, we condition \nameMethod on a fixed camera pose while varying only the temporal control signal. Experiments are performed on the withheld \nameDataset test split, which contains 50 scenes rendered with dense full-grid trajectories that can be retimed into arbitrary temporal sequences.
For each test case, we take a moving-camera source video but set the target camera trajectory to the first-frame pose. 
We then apply a range of temporal control signals, including reverse, bullet-time, zigzag, slow motion, and normal playback, to synthesize the corresponding retimed outputs. 
Since we have ground-truth frames for all temporal configurations, we report perceptual losses: PSNR, SSIM, and LPIPS.

We consider two baselines:
(1) \textit{ReCamM+preshuffled}: original \recammaster combined with input re-shuffling; and
(2) \textit{ReCamM+jointdata}: following \cite{watson2025controllingspacetimediffusion,Wu2024}, we train \recammaster with additional static-scene datasets \cite{zhou2018stereo,li2019mannequin} which provide only one single temporal pattern.

While frame shuffling may succeed in simple scenarios, it fails to disentangle camera and temporal control.
As shown in Table \ref{tab:temporal_eval}, this approach exhibits the weakest temporal controllability.
Although incorporating static-scene datasets improves performance, particularly in the bullet-time category, relying on a single temporal control pattern remains insufficient for achieving robust temporal consistency.
In contrast, \nameMethod consistently outperforms all baselines across all temporal configurations.

\subsubsection{Visual Quality Evaluation.}
Next, we evaluate the perceptual realism of our 1800 generated videos using VBench \cite{huang2023vbench}. We report all standard visual quality metrics to provide a comprehensive assessment of generative fidelity. 
\Cref{tab:vbench_visual_only} shows that our model achieves visual quality comparable to the baselines.

\begin{table}[t]
    \centering
    \caption{
        VBench visual-quality evaluation across six dimensions.
        Higher is better for all metrics. 
    }
    \vspace{-0.2cm}
    \label{tab:vbench_visual_only}
    \resizebox{\columnwidth}{!}{
    \setlength\tabcolsep{2pt}
    \begin{tabular}{lcccccc}
        \toprule
        \textbf{Method} 
        & ImgQ$\uparrow$ 
        & BGCons$\uparrow$ 
        & Motion$\uparrow$ 
        & SubjCons$\uparrow$ 
        & Flicker$\uparrow$
        & Aesthetic$\uparrow$ \\
        \midrule

        Traj-Crafter \cite{YU2025}
        & 0.6389 & \textbf{0.9376} & 0.9888 & \textbf{0.9463} & 0.9816 & 0.5172 \\

        ReCamM \cite{Bai2025}
        & 0.6302 & 0.9114 & 0.9945 & 0.9181 & \textbf{0.9825} & 0.5332 \\

        ReCamM+Aug
        & 0.6315 & 0.9165 & 0.9946 & 0.9313 & 0.9788 & 0.5385 \\

        \textbf{STPilot (Ours)}
        & \textbf{0.6486} & \textbf{0.9199} & \textbf{0.9947} & \textbf{0.9325} & 0.9781 & \textbf{0.5315} \\
        
        \bottomrule
    \end{tabular}}
\end{table}

\begin{table*}[t]
\centering
\begin{tabular}{cc}

\begin{minipage}{0.55\textwidth}
\centering
\setlength{\tabcolsep}{0.55pt}
{\footnotesize
\caption{Camera accuracy and first-frame estimation. For camera control, the enhanced camera control mechanism enables the generated video to start from an arbitrary camera angle while maintaining good camera accuracy.}
\vspace{-0.2cm}
\label{tab:camera_accuracy}

\begin{tabular}{l|cc|cc||c|cc}
\toprule
\textbf{Method} 
& RelRot$\downarrow$ & RelTrans$\downarrow$ 
& AbsRot$\downarrow$ & AbsTrans$\downarrow$ 
& Rot$^\dagger\downarrow$ 
& RTA15$^\dagger\uparrow$ & RTA30$^\dagger\uparrow$ \\
\midrule
Traj-Crafter \cite{YU2025}  & 5.94 & 0.50 & 6.93 & 0.52 & 9.76 & 22.96\% & 25.93\% \\
ReCamM \cite{Bai2025}      & 4.26 & \textbf{0.32} & 10.08 & 0.34 & 7.49 & 7.61\% & 10.20\% \\
ReCamM+Aug                 & 3.66 & 0.43 & 11.74 & 0.46 & 13.88 & 3.89\% & 5.93\% \\
\textbf{SpaceTimePilot (ours)} 
                           & \textbf{2.71} & 0.33 
                           & \textbf{5.63} & \textbf{0.34} 
                           & \textbf{4.09} 
                           & \textbf{35.19\%} & \textbf{54.44\%} \\
\bottomrule
\end{tabular}
}

\end{minipage}
&
\begin{minipage}{0.43\textwidth}
\centering
\setlength{\tabcolsep}{0.43pt}
{\footnotesize
\caption{Time-embedding compressor ablation. The proposed time-embedding method, trained with temporal warping on the proposed dataset, yields sharper results overall.}
\vspace{-0.2cm}
\label{tab:time_embed_ablation}

\begin{tabular}{lccc}
\toprule
Time Embedding & PSNR$\uparrow$ & SSIM$\uparrow$ & LPIPS$\downarrow$ \\
\midrule
Uniform Sampling     & 14.10 & 0.5981 & 0.5039 \\
1D-Conv              & 14.75 & 0.6134 & 0.4878 \\
1D-Conv + Joint Data & 15.41 & 0.6252 & 0.4830 \\
\textbf{1D-Conv {\small+\nameDataset}} 
                     & \textbf{21.16} & \textbf{0.7674} & \textbf{0.1764} \\
\bottomrule
\end{tabular}
}
\end{minipage}

\end{tabular}

{\footnotesize
\noindent\makebox[\linewidth][l]{$^\dagger$~Evaluation based on first-frame camera accuracy.}
}
\end{table*}

\subsubsection{Camera-Control Evaluation.}
Finlay, we evaluate the effectiveness of our camera control mechanism detailed in Sec.~\ref{sec:src_cam}.
Unlike the retiming evaluation above, which relies on synthetic ground-truth videos, here we construct a real-world 90-video evaluation set from OpenVideoHD~\cite{nan2024openvid}, encompassing diverse dynamic human and object motions.
Each method is evaluated across 20 camera trajectories: 10 starting from the same initial pose as the source video and 10 from different initial poses, resulting in a total of 1800 generated videos.
We apply SpatialTracker-v2~\cite{Xiao2025} to recover camera poses from the generated videos and compare them with the corresponding input camera poses.
\todo{To ensure consistent scale, we align the magnitude of the first two camera locations.}
Trajectory accuracy is quantified using $\mathbf{RotErr}$ and $\mathbf{TransErr}$ following~\cite{He2025camctrl}, under two protocols: (1) evaluating the raw trajectories defined w.r.t.~the first frame (relative protocol, RelRot, RelTrans) and (2) evaluating after aligning to the estimated pose of the first frame (absolute protocol, AbsRot, AbsTrans).
Specifically, we transform the recovered raw trajectories by multiplying the relative pose between the generated and source first frames, estimated by \duster~\cite{dust3r_cvpr24}. 
We also compare this \duster pose with the target trajectory’s initial pose, and report RotErr, RTA@15 and RTA@30, as translation magnitude is scale-ambiguous.

\begin{figure}[t]
    \centering
    \vspace{-3mm}
    \includegraphics[width=\linewidth,trim={0 0 0 0},clip]{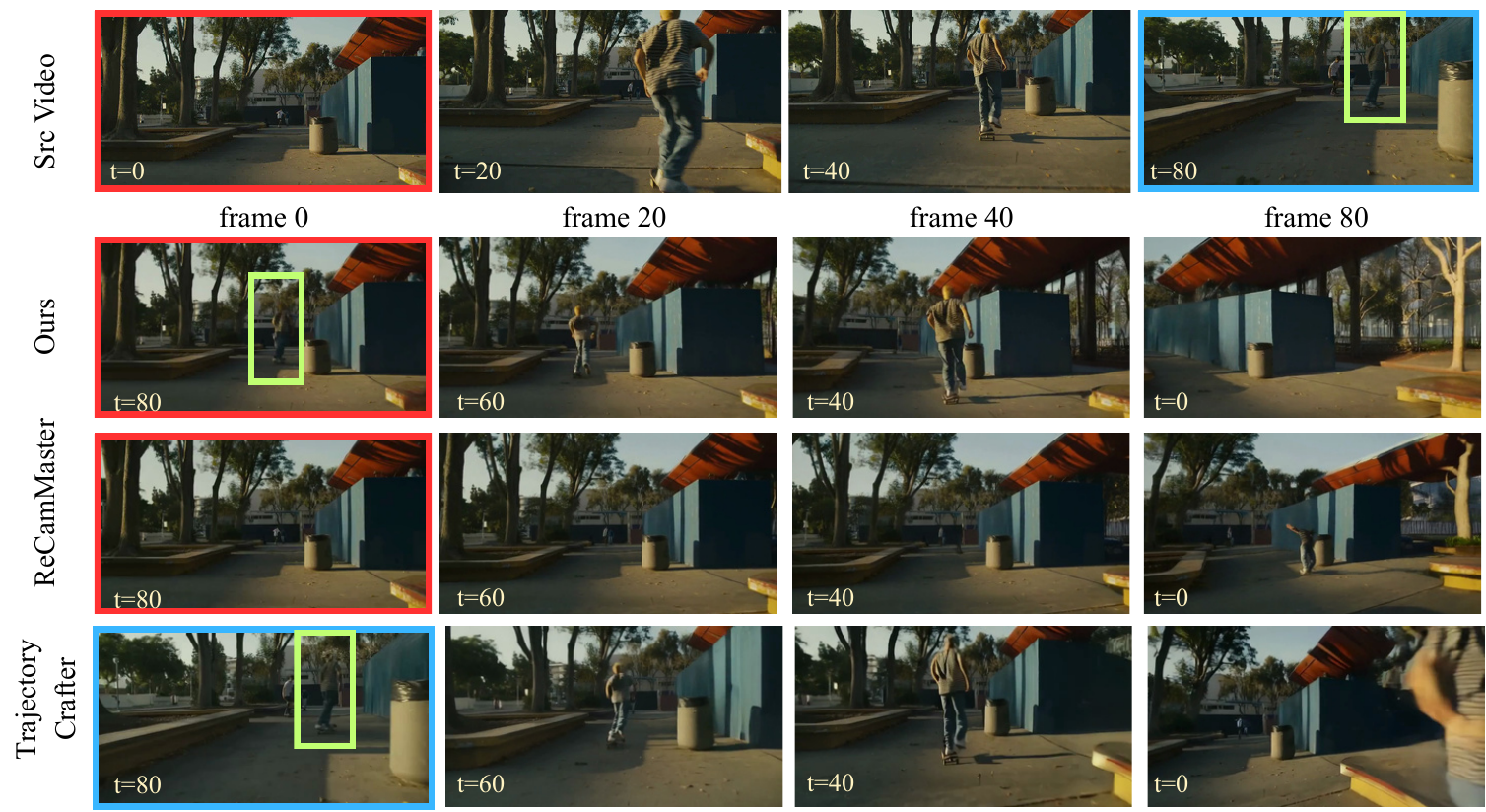}
    \caption{
    \textbf{Qualitative comparison of disentangled camera-time control.}  In this example, we apply reverse playback (time) and a pan-right camera motion starting from the first-frame pose to a source video (top), whose original camera motion is dolly-in (red to blue). SpaceTimePilot, by explicitly disentangling space and time, achieves correct camera control (red boxes) together with accurate temporal control (green boxes). For TrajectoryCrafter, it  first reverses the frames and then apply their method for viewpoint control, resulting in incorrect  camera motion. ReCamMaster (with joint-dataset training) is unable to perform temporal control, leading to failure cases.}
    \label{fig:disentangle}
\end{figure}

\begin{figure}[t]
    \centering
    \includegraphics[width=1\linewidth, trim={5mm 5mm 5mm 1mm}, clip]{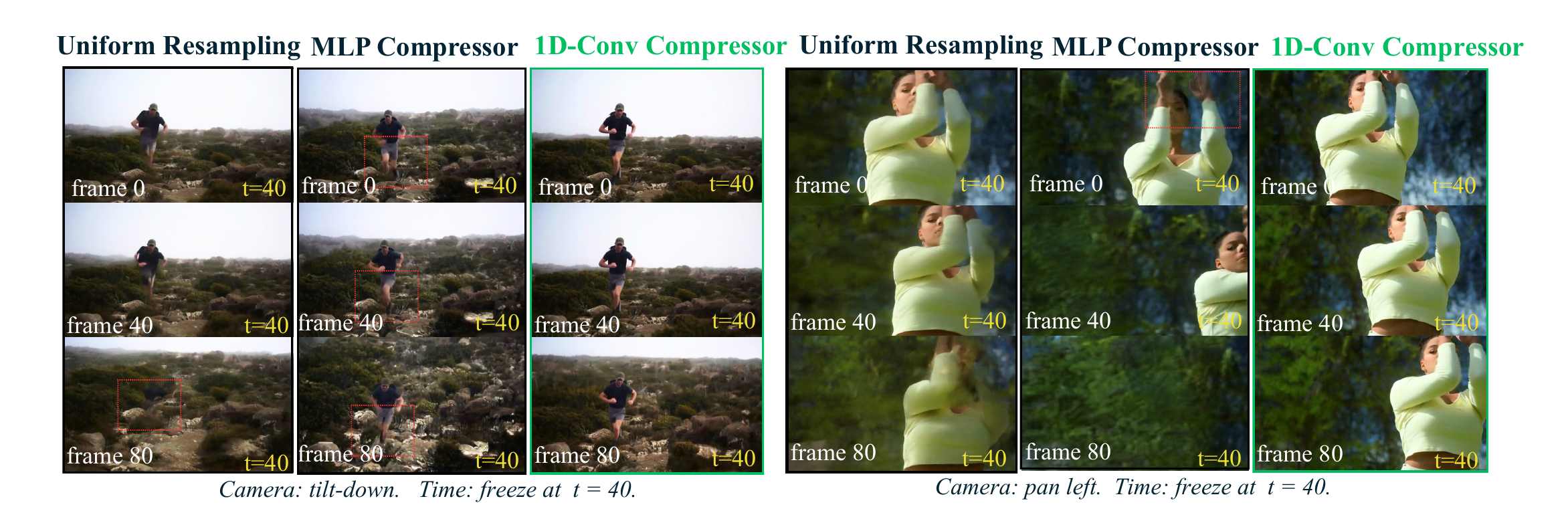}
    \caption{
        \textbf{Temporal compression ablation.}
        Comparing uniform resampling, MLP, and 1D-Conv compressors under tilt-down and pan-right bullet-time controls, $\animatime_\trg = [40,\dots,40]$.
    }
    \label{fig:temporal_ablation}
\end{figure}

To measure only the impact of source camera conditioning, we consider the original \recammaster~\cite{Bai2025} (\textit{ReCamM}) and two variants. 
Since \recammaster is originally trained on datasets where the first frame of the source and target videos are identical, the model always copies the first frame regardless of the input camera pose. 
For fairness, we retrain \recammaster with more data augmentations to include non-identical first frames, denoted as \textit{ReCamM+Aug}. 
Next, we condition the model additionally with source cameras $\camera_\src$ following Eq.~\ref{eq-srccam}, denoted as \textit{ReCamM+Aug+$\camera_\src$}. 
Finally we also report the results of TrajectoryCrafter~\cite{YU2025}.

In Table~\ref{tab:camera_accuracy}, we observe that the absolute protocol produces consistently higher errors, as trajectories must not only match the overall shape (relative protocol) but also align correctly in position and orientation.
Interestingly, ReCamM+Aug yields higher errors than the original ReCamM, whereas incorporating source cameras $\camera_\src$ results in the best overall performance.
This suggests that, without explicit reference to $\camera_\src$, exposure to more augmented videos with differing initial frames can instead confuse the model.
The newly introduced conditioning signal on the source video’s trajectory $\camera_\src$ achieves substantially better camera-control accuracy across all metrics, more reliable first-frame alignment, and more faithful adherence to the full trajectory than all baselines. 

\subsubsection{Qualitative results.}
Besides the quantitative evaluation, we also demonstrate the strength of \nameMethod with visual examples.
In \cref{fig:disentangle}, we show that only our method correctly synthesizes both the camera motion (red boxes) and the animation-time state (green boxes). While \recammaster handles camera control well, it cannot modify the temporal state, such as enabling reverse playback. TrajectoryCrafter, in contrast, is confused by the reverse frame shuffle, causing the camera pose of the last source frame (blue boxes) to incorrectly appear in the first frame of the generated video. More visual results can be found in \cref{fig:quali}.
\subsection{Ablation Study}

\label{sec:time_embedding_ablation}
To validate the effectiveness of the proposed Time embedding module, in Table \ref{tab:time_embed_ablation}, we follow the time-control evaluation set up in Sec.~\ref{sec:eval:time_control} and compare our 1D convolutional time embedding against several variants and alternatives discussed in Sec.~\ref{sec:time_embedding}:
(1) Uniform-Sampling: sampling the 81-frame embedding uniformly to a 21-frame sequence, which is equivalent to adopting sinusoidal embeddings at the latent frame $f'$ level;
(2) 1D-Conv: using 1D convolution layers to compress from $\animatime \in \mathbb{R}^F$ to $\animatime \in \mathbb{R}^{F'}$, trained with ReCamMaster and SynCamMaster datasets.
(3) 1D-Conv+jointdata: row 2 but including additionally static-scene datasets \cite{zhou2018stereo,li2019mannequin}.
(4) 1D-Conv (ours): row 2 but instead including the proposed \nameDataset.  
We observe that applying a 1D convolution to learn a compact representation by compressing the fine-grained $\numf$-dim embeddings into a $\numflt$-dim space performs notably better than directly constructing sinusoidal embeddings at the coarse $f'$ level.
Incorporating static-scene datasets yields only limited improvements, likely due to their restricted temporal control patterns.
By contrast, using the proposed \nameDataset consistently delivers the largest gains across all three metrics, confirming the effectiveness of our newly introduced datasets.
Furthermore, as shown in \cref{fig:temporal_ablation}, we present a visual comparison of bullet-time results using uniform sampling and an MLP instead of the 1D convolution for compressing the temporal control signal.
Uniform sampling produces noticeable artifacts, and the MLP compressor causes abrupt camera motion, whereas the 1D convolution effectively locks the animation time and enables smooth camera movement.

\section{Conclusion}
We present \nameMethod, the first video diffusion model to provide fully disentangled spatial and temporal control, enabling 4D space-time exploration from a single monocular video.
Our method introduces a new “animation time” representation together with a source-aware camera-control mechanism that leverages both source and target poses. This is supported by the synthetic \nameDataset and a temporal-warping training scheme, which supply dense spatiotemporal supervision.
These components allow precise camera and time manipulation, arbitrary initial poses, and flexible multi-round generation.
Across extensive experiments, \nameMethod consistently surpasses state-of-the-art baselines, offering significantly improved camera-control accuracy and reliable execution of complex retiming effects such as reverse playback, slow motion, and bullet-time. 

\section{Acknowledgement}

We would like to extend our gratitude to Duygu Ceylan, Paul Guerrero, and Zifan Shi for insightful discussions and valuable feedback on the manuscript. We also thank Rudi Wu for helpful discussions on implementation details of CAT4D.

{
    \small
    \bibliographystyle{ieeenat_fullname}
    \bibliography{main}

@String(IJCV = {Int. J. Comput. Vis.})

@String(CVPR= {IEEE Conf. Comput. Vis. Pattern Recog.})

@String(ICCV= {Int. Conf. Comput. Vis.})

@String(ECCV= {Eur. Conf. Comput. Vis.})

@String(NeurIPS= {Adv. Neural Inform. Process. Syst.})

@String(TOG= {ACM Trans. Graph.})

@String(ICLR = {Int. Conf. Learn. Represent.})

@String(IJCV  = {IJCV})

@String(CVPR  = {CVPR})

@String(ICCV  = {ICCV})

@String(ECCV  = {ECCV})

@String(NIPS  = {NeurIPS})

@String(TOG   = {ACM TOG})

@String(ICLR  = {ICLR})

@inproceedings{Wu2024,
   author = {Rundi Wu and Ruiqi Gao and Ben Poole and Alex Trevithick and Changxi Zheng and Jonathan T Barron and Aleksander Holynski},
   booktitle =CVPR,
   title = {{Cat4D}: Create anything in 4d with multi-view video diffusion models},
   year = {2024}
}

@article{wu2025video,
  title={Video World Models with Long-term Spatial Memory},
  author={Wu, Tong and Yang, Shuai and Po, Ryan and Xu, Yinghao and Liu, Ziwei and Lin, Dahua and Wetzstein, Gordon},
  journal={arXiv preprint arXiv:2506.05284},
  year={2025}
}

@inproceedings{Bai2025,
   author = {Jianhong Bai and Menghan Xia and Xiao Fu and Xintao Wang and Lianrui Mu and Jinwen Cao and Zuozhu Liu and Haoji Hu and Xiang Bai and Pengfei Wan and Di Zhang},
   booktitle =ICCV,
   title = {{ReCamMaster}: Camera-Controlled Generative Rendering from A Single Video},
   year = {2025}
}

@article{He2025,
   author = {Hao He and Ceyuan Yang and Shanchuan Lin and Yinghao Xu and Meng Wei and Liangke Gui and Qi Zhao and Gordon Wetzstein and Lu Jiang and Hongsheng Li},
   journal = {arXiv preprint arXiv:2503.10592},
   title = {{CameraCtrl II}: Dynamic Scene Exploration via Camera-controlled Video Diffusion Models},
   year = {2025}
}

@inproceedings{Ren2025,
   author = {Xuanchi Ren and Tianchang Shen and Jiahui Huang and Huan Ling and Yifan Lu and Merlin Nimier-David and Thomas M\"uller and Alexander Keller and Sanja Fidler and Jun Gao},
   booktitle = CVPR,
   title = {{GEN3C}: 3D-Informed World-Consistent Video Generation with Precise Camera Control},
   year = {2025}
}

@inproceedings{Jeong2025,
   author = {Hyeonho Jeong and Suhyeon Lee and Jong Chul Ye},
   booktitle = ICCV,
   title = {{Reangle-A-Video}: 4D Video Generation as Video-to-Video Translation},
   year = {2025}
}

@inproceedings{YU2025,
   author = {Mark YU and Wenbo Hu and Jinbo Xing and Ying Shan},
   booktitle = ICCV,
   title = {{TrajectoryCrafter}: Redirecting Camera Trajectory for Monocular Videos via Diffusion Models},
   year = {2025}
}

@inproceedings{Zhang2024,
   author = {David Junhao Zhang and Roni Paiss and Shiran Zada and Nikhil Karnad and David E Jacobs and Yael Pritch and Inbar Mosseri and Mike Zheng Shou and Neal Wadhwa and Nataniel Ruiz},
   booktitle = CVPR,
   title = {{ReCapture}: Generative Video Camera Controls for User-Provided Videos using Masked Video Fine-Tuning},
   year = {2024}
}

@article{Zhou2025,
   author = {Jensen (Jinghao) Zhou and Hang Gao and Vikram Voleti and Aaryaman Vasishta and Chun-Han Yao and Mark Boss and Philip Torr and Christian Rupprecht and Varun Jampani},
   journal = {arXiv preprint},
   title = {{Stable Virtual Camera}: Generative View Synthesis with Diffusion Models},
   year = {2025}
}

@inproceedings{jin2025lvsm,
    title={{LVSM}: A Large View Synthesis Model with Minimal 3D Inductive Bias},
    author={Haian Jin and Hanwen Jiang and Hao Tan and Kai Zhang and Sai Bi and Tianyuan Zhang and Fujun Luan and Noah Snavely and Zexiang Xu},
    booktitle=ICLR,
    year={2025},
    url={https://openreview.net/forum?id=QQBPWtvtcn}
}

@inproceedings{He2025camctrl,
   author = {Hao He and Yinghao Xu and Yuwei Guo and Gordon Wetzstein and Bo Dai and Hongsheng Li and Ceyuan Yang},
   booktitle = ICLR,
   title = {{CameraCtrl}: Enabling Camera Control for Text-to-Video Generation},
   year = {2025}
}

@inproceedings{Xiao2025,
   author = {Yuxi Xiao and Jianyuan Wang and Nan Xue and Nikita Karaev and Iurii Makarov and Bingyi Kang and Xin Zhu and Hujun Bao and Yujun Shen and Xiaowei Zhou},
   booktitle = ICCV,
   title = {{SpatialTrackerV2}: 3D Point Tracking Made Easy},
   year = {2025}
}

@article{zhang2024show,
  title={Show-1: Marrying pixel and latent diffusion models for text-to-video generation},
  author={Zhang, David Junhao and Wu, Jay Zhangjie and Liu, Jia-Wei and Zhao, Rui and Ran, Lingmin and Gu, Yuchao and Gao, Difei and Shou, Mike Zheng},
  journal=IJCV,
  pages={1--15},
  year={2024},
  publisher={Springer}
}

@article{wang2024lavie,
  title={Lavie: High-quality video generation with cascaded latent diffusion models},
  author={Wang, Yaohui and Chen, Xinyuan and Ma, Xin and Zhou, Shangchen and Huang, Ziqi and Wang, Yi and Yang, Ceyuan and He, Yinan and Yu, Jiashuo and Yang, Peiqing and others},
  journal=IJCV,
  pages={1--20},
  year={2024},
  publisher={Springer}
}

@inproceedings{jeong2024track4gen,
  title={{Track4Gen}: Teaching Video Diffusion Models to Track Points Improves Video Generation},
  author={Jeong, Hyeonho and Huang, Chun-Hao Paul and Ye, Jong Chul and Mitra, Niloy and Ceylan, Duygu},
  booktitle=CVPR,
  year={2025}
}

@inproceedings{bar2024lumiere,
  title={Lumiere: A space-time diffusion model for video generation},
  author={Bar-Tal, Omer and Chefer, Hila and Tov, Omer and Herrmann, Charles and Paiss, Roni and Zada, Shiran and Ephrat, Ariel and Hur, Junhwa and Liu, Guanghui and Raj, Amit and others},
  booktitle={SIGGRAPH Asia 2024 Conference Papers},
  pages={1--11},
  year={2024}
}

@misc{polyak2024moviegencastmedia,
      title={Movie Gen: A Cast of Media Foundation Models}, 
      author={Adam Polyak and Amit Zohar and Andrew Brown and Andros Tjandra and Animesh Sinha and Ann Lee and Apoorv Vyas and Bowen Shi and Chih-Yao Ma and Ching-Yao Chuang and David Yan and Dhruv Choudhary and Dingkang Wang and Geet Sethi and Guan Pang and Haoyu Ma and Ishan Misra and Ji Hou and Jialiang Wang and Kiran Jagadeesh and Kunpeng Li and Luxin Zhang and Mannat Singh and Mary Williamson and Matt Le and Matthew Yu and Mitesh Kumar Singh and Peizhao Zhang and Peter Vajda and Quentin Duval and Rohit Girdhar and Roshan Sumbaly and Sai Saketh Rambhatla and Sam Tsai and Samaneh Azadi and Samyak Datta and Sanyuan Chen and Sean Bell and Sharadh Ramaswamy and Shelly Sheynin and Siddharth Bhattacharya and Simran Motwani and Tao Xu and Tianhe Li and Tingbo Hou and Wei-Ning Hsu and Xi Yin and Xiaoliang Dai and Yaniv Taigman and Yaqiao Luo and Yen-Cheng Liu and Yi-Chiao Wu and Yue Zhao and Yuval Kirstain and Zecheng He and Zijian He and Albert Pumarola and Ali Thabet and Artsiom Sanakoyeu and Arun Mallya and Baishan Guo and Boris Araya and Breena Kerr and Carleigh Wood and Ce Liu and Cen Peng and Dimitry Vengertsev and Edgar Schonfeld and Elliot Blanchard and Felix Juefei-Xu and Fraylie Nord and Jeff Liang and John Hoffman and Jonas Kohler and Kaolin Fire and Karthik Sivakumar and Lawrence Chen and Licheng Yu and Luya Gao and Markos Georgopoulos and Rashel Moritz and Sara K. Sampson and Shikai Li and Simone Parmeggiani and Steve Fine and Tara Fowler and Vladan Petrovic and Yuming Du},
      year={2024},
      eprint={2410.13720},
      archivePrefix={arXiv},
      primaryClass={cs.CV},
      url={https://arxiv.org/abs/2410.13720}, 
}

@article{videoworldsimulators2024,
  title={Video generation models as world simulators},
  author={Tim Brooks and Bill Peebles and Connor Holmes and Will DePue and Yufei Guo and Li Jing and David Schnurr and Joe Taylor and Troy Luhman and Eric Luhman and Clarence Ng and Ricky Wang and Aditya Ramesh},
  year={2024},
  url={https://openai.com/research/video-generation-models-as-world-simulators},
}

@article{kong2024hunyuanvideo,
  title={Hunyuanvideo: A systematic framework for large video generative models},
  author={Kong, Weijie and Tian, Qi and Zhang, Zijian and Min, Rox and Dai, Zuozhuo and Zhou, Jin and Xiong, Jiangfeng and Li, Xin and Wu, Bo and Zhang, Jianwei and others},
  journal={arXiv preprint arXiv:2412.03603},
  year={2024}
}

@article{yang2024cogvideox,
  title={Cogvideox: Text-to-video diffusion models with an expert transformer},
  author={Yang, Zhuoyi and Teng, Jiayan and Zheng, Wendi and Ding, Ming and Huang, Shiyu and Xu, Jiazheng and Yang, Yuanming and Hong, Wenyi and Zhang, Xiaohan and Feng, Guanyu and others},
  journal={arXiv preprint arXiv:2408.06072},
  year={2024}
}

@article{blattmann2023stable,
  title={Stable video diffusion: Scaling latent video diffusion models to large datasets},
  author={Blattmann, Andreas and Dockhorn, Tim and Kulal, Sumith and Mendelevitch, Daniel and Kilian, Maciej and Lorenz, Dominik and Levi, Yam and English, Zion and Voleti, Vikram and Letts, Adam and others},
  journal={arXiv preprint arXiv:2311.15127},
  year={2023}
}

@article{wan2025,
      title={Wan: Open and Advanced Large-Scale Video Generative Models}, 
      author={Team Wan and Ang Wang and Baole Ai and Bin Wen and Chaojie Mao and Chen-Wei Xie and Di Chen and Feiwu Yu and Haiming Zhao and Jianxiao Yang and Jianyuan Zeng and Jiayu Wang and Jingfeng Zhang and Jingren Zhou and Jinkai Wang and Jixuan Chen and Kai Zhu and Kang Zhao and Keyu Yan and Lianghua Huang and Mengyang Feng and Ningyi Zhang and Pandeng Li and Pingyu Wu and Ruihang Chu and Ruili Feng and Shiwei Zhang and Siyang Sun and Tao Fang and Tianxing Wang and Tianyi Gui and Tingyu Weng and Tong Shen and Wei Lin and Wei Wang and Wei Wang and Wenmeng Zhou and Wente Wang and Wenting Shen and Wenyuan Yu and Xianzhong Shi and Xiaoming Huang and Xin Xu and Yan Kou and Yangyu Lv and Yifei Li and Yijing Liu and Yiming Wang and Yingya Zhang and Yitong Huang and Yong Li and You Wu and Yu Liu and Yulin Pan and Yun Zheng and Yuntao Hong and Yupeng Shi and Yutong Feng and Zeyinzi Jiang and Zhen Han and Zhi-Fan Wu and Ziyu Liu},
      journal = {arXiv preprint arXiv:2503.20314},
      year={2025}
}

@inproceedings{liang2024diffusion4d,
  title={{Diffusion4D}: Fast Spatial-temporal Consistent
    4D Generation via Video Diffusion Models},
  author={Liang, Hanwen and Yin, Yuyang and Xu, Dejia and Liang, Hanxue and Wang, Zhangyang and Plataniotis, Konstantinos N and Zhao, Yao and Wei, Yunchao},
  booktitle = NIPS,
  year={2024}
}

@article{vanhoorick2024gcd,
    title={Generative Camera Dolly: Extreme Monocular Dynamic Novel View Synthesis},
    author={Van Hoorick, Basile and Wu, Rundi and Ozguroglu, Ege and Sargent, Kyle and Liu, Ruoshi and Tokmakov, Pavel and Dave, Achal and Zheng, Changxi and Vondrick, Carl},
    journal={ECCV},
    year={2024}
}

@article{zun2025epic,
				author    = {Wang, Zun and Cho, Jaemin and Li, Jialu and Lin, Han and Yoon, Jaehong and Zhang, Yue and Bansal, Mohit},
				title     = {{EPiC: Efficient Video Camera Control Learning with Precise Anchor-Video}},
				journal   = {arXiv preprint arXiv:2505.21876},
				year      = {2025},
				url       = {http://arxiv.org/abs/2505.21876}
			  }

@inproceedings{greff2021kubric,
    title = {Kubric: a scalable dataset generator}, 
    author = {Klaus Greff and Francois Belletti and Lucas Beyer and Carl Doersch and
              Yilun Du and Daniel Duckworth and David J Fleet and Dan Gnanapragasam and
              Florian Golemo and Charles Herrmann and Thomas Kipf and Abhijit Kundu and
              Dmitry Lagun and Issam Laradji and Hsueh-Ti (Derek) Liu and Henning Meyer and
              Yishu Miao and Derek Nowrouzezahrai and Cengiz Oztireli and Etienne Pot and
              Noha Radwan and Daniel Rebain and Sara Sabour and Mehdi S. M. Sajjadi and Matan Sela and
              Vincent Sitzmann and Austin Stone and Deqing Sun and Suhani Vora and Ziyu Wang and
              Tianhao Wu and Kwang Moo Yi and Fangcheng Zhong and Andrea Tagliasacchi},
    booktitle = CVPR,
    year = {2022},
}

@inproceedings{li2023dynibar,
  title         = {DynIBaR: Neural Dynamic Image-Based Rendering},
  author        = {Li, Zhengqi and Wang, Qianqian and Cole, Forrester and Tucker, Richard and Snavely, Noah},
  booktitle     = {CVPR},
  year          = {2023},
}

@article{yu2024viewcrafter,
    title={ViewCrafter: Taming Video Diffusion Models for High-fidelity Novel View Synthesis},
    author={Yu, Wangbo and Xing, Jinbo and Yuan, Li and Hu, Wenbo and Li, Xiaoyu and Huang, Zhipeng and Gao, Xiangjun and Wong, Tien-Tsin and Shan, Ying and Tian, Yonghong},
    journal={arXiv preprint arXiv:2409.02048},
    year={2024}
  }

@inproceedings{watson2025controllingspacetimediffusion,
      title={Controlling Space and Time with Diffusion Models}, 
      author={Daniel Watson and Saurabh Saxena and Lala Li and Andrea Tagliasacchi and David J. Fleet},
      year={2025},
        booktitle = ICLR,
      eprint={2407.07860},
      archivePrefix={arXiv},
      primaryClass={cs.CV},
      url={https://arxiv.org/abs/2407.07860}, 
}

@article{bai2024syncammaster,
  title={{SynCamMaster}: Synchronizing Multi-Camera Video Generation from Diverse Viewpoints},
  author={Bai, Jianhong and Xia, Menghan and Wang, Xintao and Yuan, Ziyang and Fu, Xiao and Liu, Zuozhu and Hu, Haoji and Wan, Pengfei and Zhang, Di},
  journal={arXiv preprint arXiv:2412.07760},
  year={2024}
}

@inproceedings{park2021nerfies,
  title     = {Nerfies: Deformable Neural Radiance Fields},
  author    = {Park, Keunhong and Sinha, Utkarsh and Barron, Jonathan T. and Bouaziz, Sofien and Goldman, Dan B. and Seitz, Steven M. and Martin-Brualla, Ricardo},
  booktitle = ICCV,
  year      = {2021},
  pages     = {5865--5874}
}

@article{park2021hypernerf,
  title     = {{HyperNeRF}: A Higher-Dimensional Representation for Topologically Varying Neural Radiance Fields},
  author    = {Park, Keunhong and Sinha, Utkarsh and Hedman, Peter and Barron, Jonathan T. and Bouaziz, Sofien and Goldman, Dan B. and Martin-Brualla, Ricardo and Seitz, Steven M.},
  journal   = {ACM Transactions on Graphics (TOG)},
  volume    = {40},
  number    = {6},
  pages     = {238:1--238:12},
  year      = {2021},
  publisher = {ACM},
  doi       = {10.1145/3478513.3480480}
}

@inproceedings{gao2021dynamic,
  title     = {Dynamic View Synthesis from Dynamic Monocular Video},
  author    = {Gao, Chen and Saraf, Ayush and Kopf, Johannes and Huang, Jia-Bin},
  booktitle = ICCV,
  year      = {2021},
  pages     = {5712--5721}
}

@inproceedings{yoon2020novel,
  title     = {Novel View Synthesis of Dynamic Scenes with Globally Coherent Depths from a Monocular Camera},
  author    = {Yoon, Jae Shin and Kim, Kihwan and Gallo, Orazio and Park, Hyun Soo and Kautz, Jan},
  booktitle = CVPR,
  year      = {2020},
  pages     = {5339--5348}
}

@inproceedings{li2021nsff,
  title     = {Neural Scene Flow Fields for Space-Time View Synthesis of Dynamic Scenes},
  author    = {Li, Zhengqi and Niklaus, Simon and Snavely, Noah and Wang, Oliver},
  booktitle = CVPR,
  year      = {2021},
  pages     = {6498--6508}
}

@inproceedings{lei2024mosca,
  title     = {{MoSca}: Dynamic Gaussian Fusion from Casual Videos via 4D Motion Scaffolds},
  author    = {Lei, Jiahui and Weng, Yijia and Harley, Adam and Guibas, Leonidas and Daniilidis, Kostas},
  booktitle = CVPR,
  year      = {2025},
}

@inproceedings{wang2024shapeofmotion,
  title     = {Shape of Motion: 4D Reconstruction from a Single Video},
  author    = {Wang, Qianqian and Ye, Vickie and Gao, Hang and Austin, Jake and Li, Zhengqi and Kanazawa, Angjoo},
  booktitle = ICCV,
  year      = {2025},
}

@misc{parker-holder2025genie3,
  title        = {Genie 3: A New Frontier for World Models},
  author       = {Parker-Holder, Jack and Fruchter, Shlomi},
  howpublished = {Google DeepMind Blog},
  year         = {2025},
  month        = {Aug},
  day          = {5},
  url          = {https://deepmind.google/discover/blog/genie-3-a-new-frontier-for-world-models/},
  note         = {Accessed: <insert date you retrieved>}
}

@article{nan2024openvid,
  title={{OpenVid-1M}: A Large-Scale High-Quality Dataset for Text-to-video Generation},
  author={Nan, Kepan and Xie, Rui and Zhou, Penghao and Fan, Tiehan and Yang, Zhenheng and Chen, Zhijie and Li, Xiang and Yang, Jian and Tai, Ying},
  journal={arXiv preprint arXiv:2407.02371},
  year={2024}
}

@inproceedings{rockwell2025dynpose,
   title     = {Dynamic Camera Poses and Where to Find Them},
   author    = {Rockwell, Chris and Tung, Joseph and Lin, Tsung-Yi and Liu, Ming-Yu and Fouhey, David F. and Lin, Chen-Hsuan},
   booktitle = CVPR,
   year      = {2025}
 }

@inproceedings{ling2024dl3dv,
  title={{DL3DV-10K}: A large-scale scene dataset for deep learning-based 3d vision},
  author={Ling, Lu and Sheng, Yichen and Tu, Zhi and Zhao, Wentian and Xin, Cheng and Wan, Kun and Yu, Lantao and Guo, Qianyu and Yu, Zixun and Lu, Yawen and others},
  booktitle=CVPR,
  pages={22160--22169},
  year={2024}
}

@inproceedings{zhou2018stereo,
    Author = {Zhou, Tinghui and Tucker, Richard and Flynn, John and Fyffe, Graham and Snavely, Noah},
    Title = {Stereo Magnification: Learning View Synthesis using Multiplane Images},
    Booktitle = {SIGGRAPH},
    Year = {2018}
}

@article{lin2024dynamic,
  title={Dynamic {NeRF}: A review},
  author={Lin, Jinwei},
  journal={arXiv preprint arXiv:2405.08609},
  year={2024}
}

@inproceedings{mildenhall2021nerf,
  title={{NeRF}: Representing scenes as neural radiance fields for view synthesis},
  author={Mildenhall, Ben and Srinivasan, Pratul P and Tancik, Matthew and Barron, Jonathan T and Ramamoorthi, Ravi and Ng, Ren},
  booktitle=ECCV,
  volume={65},
  number={1},
  pages={99--106},
  year={2021},
  publisher={ACM New York, NY, USA}
}

@inproceedings{dust3r_cvpr24,
      title={{DUSt3R}: Geometric 3D Vision Made Easy}, 
      author={Shuzhe Wang and Vincent Leroy and Yohann Cabon and Boris Chidlovskii and Jerome Revaud},
      booktitle =CVPR,
      year = {2024}
}

@InProceedings{huang2023vbench,
      title={{VBench}: Comprehensive Benchmark Suite for Video Generative Models},
      author={Huang, Ziqi and He, Yinan and Yu, Jiashuo and Zhang, Fan and Si, Chenyang and Jiang, Yuming and Zhang, Yuanhan and Wu, Tianxing and Jin, Qingyang and Chanpaisit, Nattapol and Wang, Yaohui and Chen, Xinyuan and Wang, Limin and Lin, Dahua and Qiao, Yu and Liu, Ziwei},
      booktitle=CVPR,
      year={2024}
}

@inproceedings{wu20244d,
  title={4d gaussian splatting for real-time dynamic scene rendering},
  author={Wu, Guanjun and Yi, Taoran and Fang, Jiemin and Xie, Lingxi and Zhang, Xiaopeng and Wei, Wei and Liu, Wenyu and Tian, Qi and Wang, Xinggang},
  booktitle={Proceedings of the IEEE/CVF conference on computer vision and pattern recognition},
  pages={20310--20320},
  year={2024}
}

@article{kerbl20233d,
  title={3D Gaussian splatting for real-time radiance field rendering.},
  author={Kerbl, Bernhard and Kopanas, Georgios and Leimk{\"u}hler, Thomas and Drettakis, George},
  journal={ACM Trans. Graph.},
  volume={42},
  number={4},
  pages={139--1},
  year={2023}
}

@inproceedings{li2019mannequin,
  title={Learning the Depths of Moving People by Watching Frozen People},
  author={Li, Zhengqi and Dekel, Tali and Cole, Forrester and Tucker, Richard
    and Snavely, Noah and Liu, Ce and Freeman, William T},
  booktitle=CVPR,
  year={2019}
}

@InProceedings{Lu_2025_HUMOTO,
      author    = {Lu, Jiaxin and Huang, Chun-Hao Paul and Bhattacharya, Uttaran and Huang, Qixing and Zhou, Yi},
      title     = {HUMOTO: A 4D Dataset of Mocap Human Object Interactions},
      booktitle = {Proceedings of the IEEE/CVF International Conference on Computer Vision (ICCV)},
      month     = {October},
      year      = {2025},
      pages     = {10886-10897}
}

@misc{mixamo,
author = {Adobe Systems Inc.},
title = {Mixamo},
year = {2018},
url = {https://www.mixamo.com},
note = {Accessed: 2025-03-07}
}

@InProceedings{Wang_2025_CVPR,
    author    = {Wang, Chaoyang and Zhuang, Peiye and Ngo, Tuan Duc and Menapace, Willi and Siarohin, Aliaksandr and Vasilkovsky, Michael and Skorokhodov, Ivan and Tulyakov, Sergey and Wonka, Peter and Lee, Hsin-Ying},
    title     = {4Real-Video: Learning Generalizable Photo-Realistic 4D Video Diffusion},
    booktitle=CVPR,
    month     = {June},
    year      = {2025}
}

@InProceedings{wang20254realv2,
  title={4Real-Video-V2: Fused View-Time Attention and Feedforward Reconstruction for 4D Scene Generation},
  author={Wang, Chaoyang and Mirzaei, Ashkan and Goel, Vidit and Menapace, Willi and Siarohin, Aliaksandr and Vinella, Avalon and Vasilkovsky, Michael and Skorokhodov, Ivan and Shakhrai, Vladislav and Korolev, Sergey and Tulyakov, Sergey and Wonka, Peter},
  booktitle=NeurIPS,
  year={2025}
}
}

\clearpage
\appendix

\section{Network Architecture}

\begin{figure}[h]
    \centering
    \includegraphics[width=\linewidth]{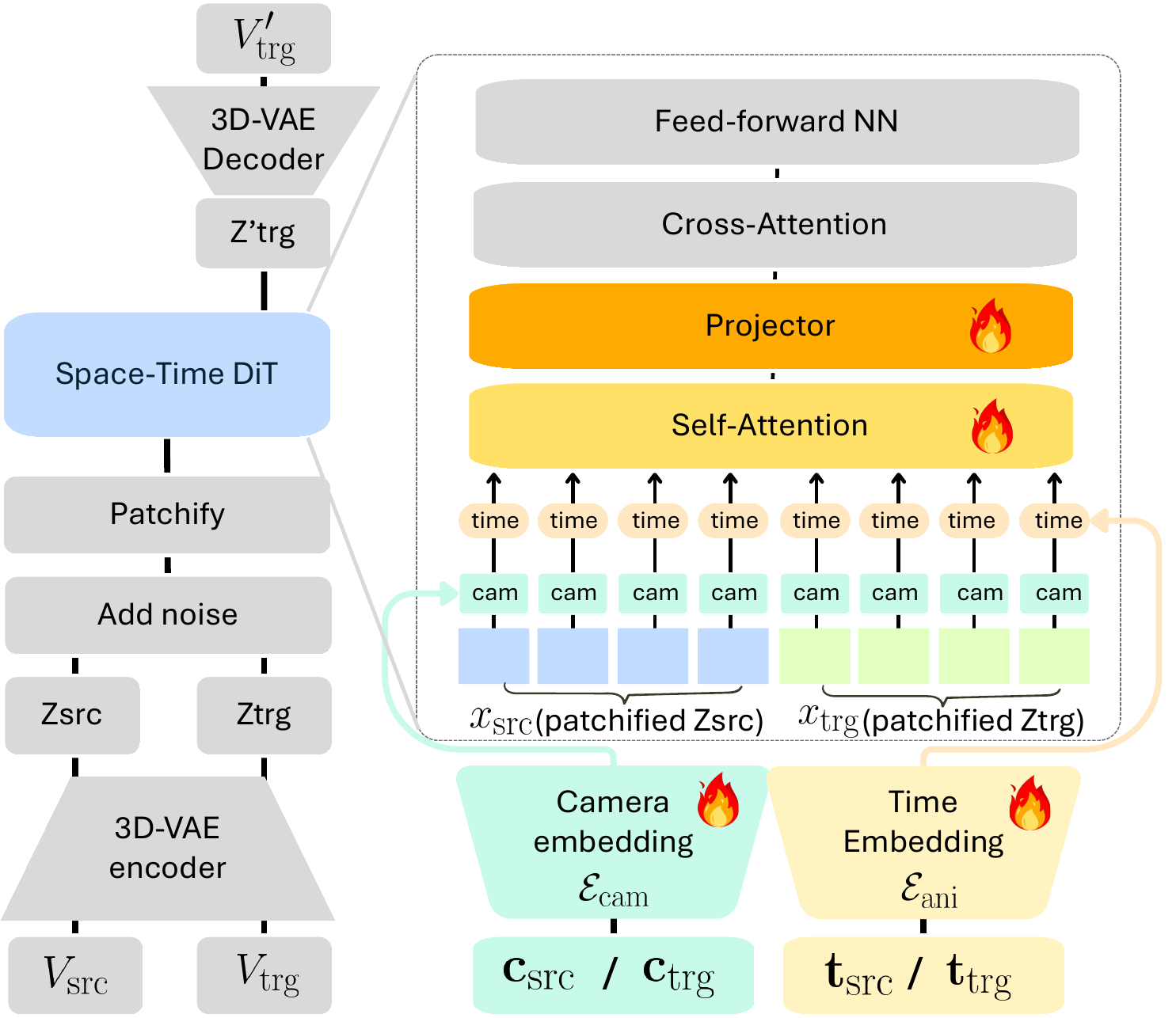}
    \caption{
        \textbf{Architecture of \nameMethod.}
        Our model jointly conditions on camera trajectories and temporal control signals via space–time attention, enabling non-monotonic motion generation such as reversals, repeats, accelerations, and zigzag time.}
    \label{fig:architecture}
\end{figure}

The network architecture of \nameMethod is depicted in \cref{fig:architecture}. 
The newly introduced animation-time embedder $\mathcal{E}_{\text{ani}}$ encodes the source and target animation times, $\animatime_\src$ and $\animatime_\trg$, into tensors matching the shapes of $x_\src$ and $x_\trg$, which are then added to them respectively.
During training, we train only the camera embedder $\mathcal{E}_{\text{cam}}$, the animation-time embedder $\mathcal{E}_{\text{ani}}$, the self-attention (full-3D attention), and the projector layers before the cross-attention.

\begin{figure*}[t]
    \centering
    \includegraphics[width=\linewidth]{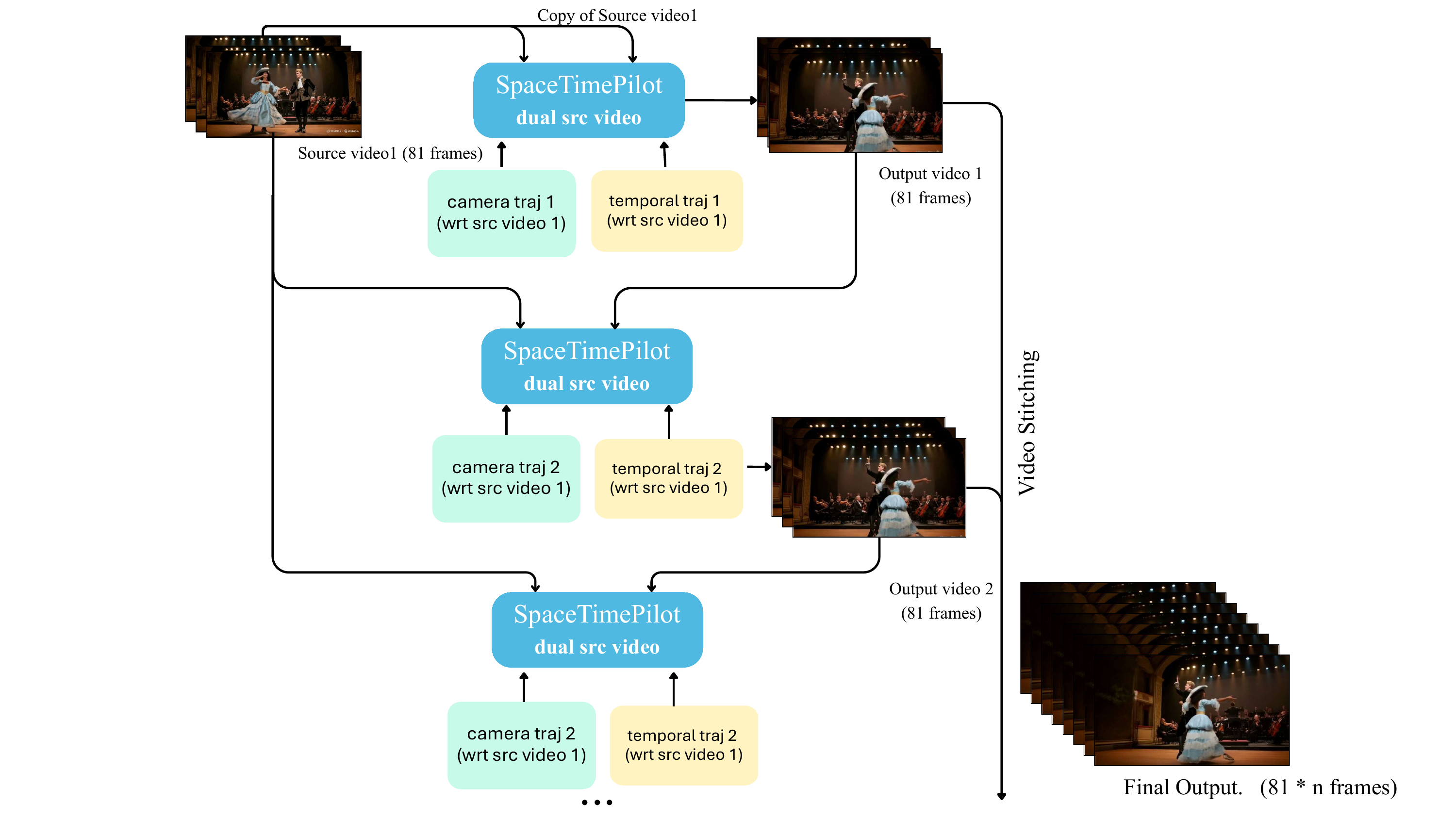}
    \caption{
        \textbf{Overview of the multi-turn autoregressive inference scheme.}
The model first generates an 81-frame segment conditioned on the source video and a chosen space–time trajectory. The resulting output is then reused as a secondary source video for subsequent iterations, each with its own camera and temporal trajectory. By chaining these iterations and stitching the outputs, \nameMethod produces a long, coherent video that follows an arbitrary space–time path.
    }
    \label{fig:multi-turn}
\end{figure*}

\section{Longer Space-Time Exploration Video with Disentangled Controls}

 One of the central advantages of \nameMethod is its ability to freely navigate both spatial and temporal dimensions, with arbitrary starting points in each dimension and fully customizable trajectories through them. Although each individual generation is limited to an 81-frame window, we show that \nameMethod can effectively extend this window indefinitely through a multi-turn autoregressive inference scheme, enabling continuous and controllable space–time exploration from a single input video. The overall pipeline is illustrated in \cref{fig:multi-turn}.

The core idea is to generate the final video in autoregressive segments that connect seamlessly.  For example, given a source video of 81 frames, we may first generate a 0.5$\times$ slow-motion sequence covering frames 0–40 with a new camera trajectory. Then, continuing both the visual context and the generated camera trajectory, we can produce the next segment starting from the final camera pose of the previous output, while temporally traversing the remaining frames 40–81. This yields an autoregressive chain of viewpoint-controlled video segments that together create a continuous long-range space–time trajectory.

A key property that enables this behavior is that our model can generate video segments whose camera poses do \emph{not} need to start at the first frame. 
This allows precise control over the starting point, both in time and viewpoint, for every generated chunk, ensuring smooth, consistent motion over extended sequences.

To maintain contextual coherence across iterations, we introduce a lightweight memory mechanism. During training, the model is conditioned on \textbf{a pair of source videos}, which enables consistent chaining during inference. Specifically:
\begin{itemize}
\item At iteration $i = 1$, the model is conditioned only on the original source video.
\item At iteration $i = 2$, it is conditioned on both the source video and the previously generated 81-frame segment.
\item This process repeats, with each iteration conditioning on the source video as well as the most recent generated segment.
\end{itemize}

This simple yet effective strategy allows \nameMethod to generate arbitrarily long, smoothly connected sequences with continuous and precise control over both temporal manipulation and camera motion.

\begin{figure*}[t]
    \vspace{-5mm}
    \centering
    \includegraphics[width=\linewidth]{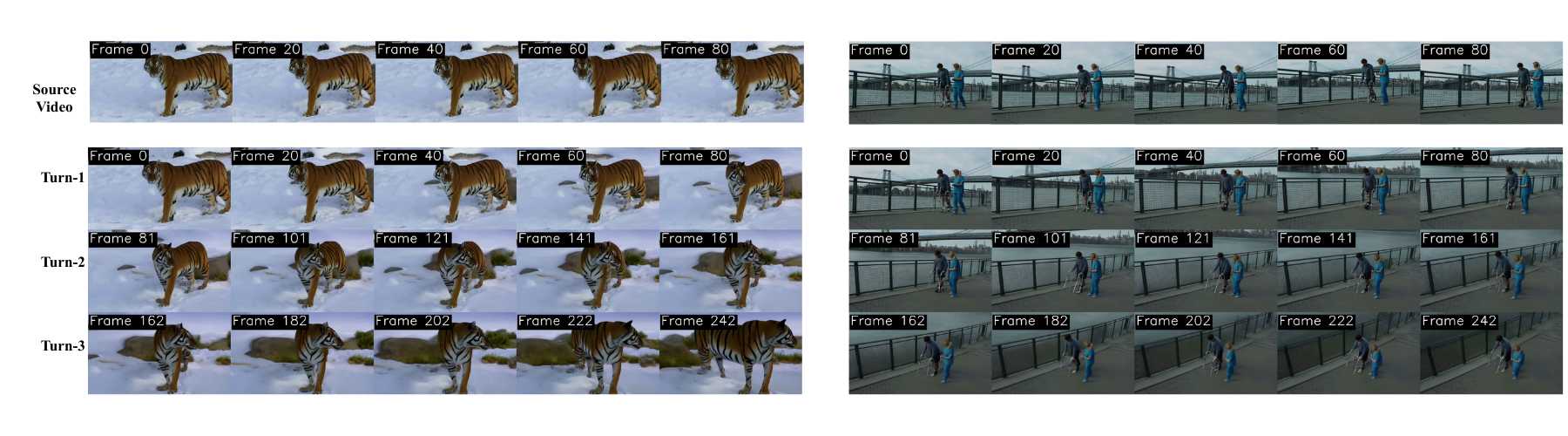}
    \vspace{-1Cm}
    \caption{
       \textbf{ Multi-turn autoregressive generation with \nameMethod.} Top row: source video frames.
Rows 2–4: Turn-1, Turn-2, and Turn-3 generations.
At each turn, \nameMethod jointly conditions on (1) the original source video and (2) the previously generated chunk, ensuring temporal continuity, stable motion progression, and consistent camera geometry.
This dual-conditioning design enables viewpoint changes far beyond the input video—such as rotating to the rear of the tiger or transitioning from a low-angle shot to a high bird’s-eye view—while preserving visual and motion coherence. Please refer to section ``AR Demos'' in the website for videos.
    }
    \label{fig:multi-turn demo}
\end{figure*}

Here, we showcase how this can be used to conduct large viewpoint changes, as demonstrated in \cref{fig:multi-turn demo}.

\section{Additional Details on the Proposed \nameDataset Dataset.}

\begin{figure}[ht]
    \centering
    \vspace{-3mm}
    \includegraphics[width=\linewidth]{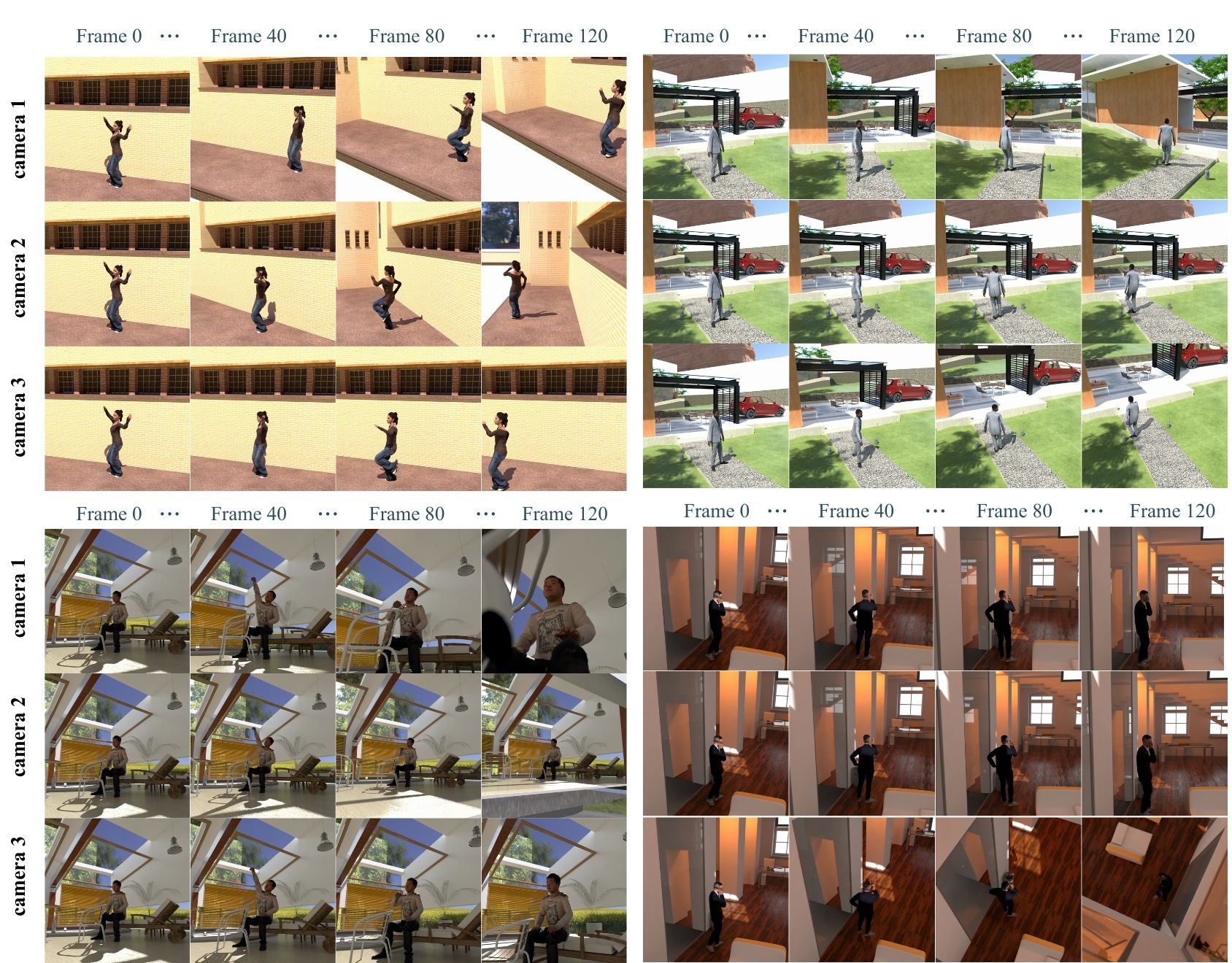}
    \caption{\textbf{Example of \nameDataset}.
   Multi-view, densely sampled sequences from the \nameDataset dataset. Each row shows frames from one camera trajectory, and each column samples different timesteps (0–120). The dataset provides diverse environments, human motions, and four camera paths per scene with full 120-frame temporal coverage.
    }
    \label{fig:visualexample}
\end{figure}

The \texttt{\nameDataset} dataset is built using high-quality, commercially licensed 3D environments that include both realistic indoor and outdoor scenes. For each environment, we populate the space with multiple animated human characters. 
The character assets are sourced from Mixamo \cite{mixamo} and HUMOTO \cite{Lu_2025_HUMOTO}, and each character is manually textured and refined to ensure realistic geometry, appearance, and material quality. 
The animations span a diverse range of human motions, including locomotion, gestures, and human-object interactions.
Examples of scenes are shown in \cref{fig:visualexample}. 
Please refer to the complementary website for the video examples.

To capture rich spatial coverage, we generate four distinct camera trajectories for every scene. Camera paths include rotational orbits, linear tracking motions, and smoothly curved arcs. A dedicated validity module ensures that each trajectory:
(1) begins at a collision-free location with clear visibility of the main character,
(2) maintains non-intersecting movement with the environment throughout the path, and
(3) preserves full subject visibility across all viewpoints.

Each trajectory is rendered into a 120-frame sequence at a resolution of $1080 \times 1080$ pixels, providing dense temporal sampling with high visual fidelity. This yields three multi-view video sequences per scene, each covering the full motion duration with consistent lighting, textures, and geometry. Overall, we rendered 1500 videos from 500 animations, each with 120 videos full grid rendering, leading to 180k videos.

\begin{figure}[ht]
    \centering
    \vspace{-3mm}
    \includegraphics[width=\linewidth]{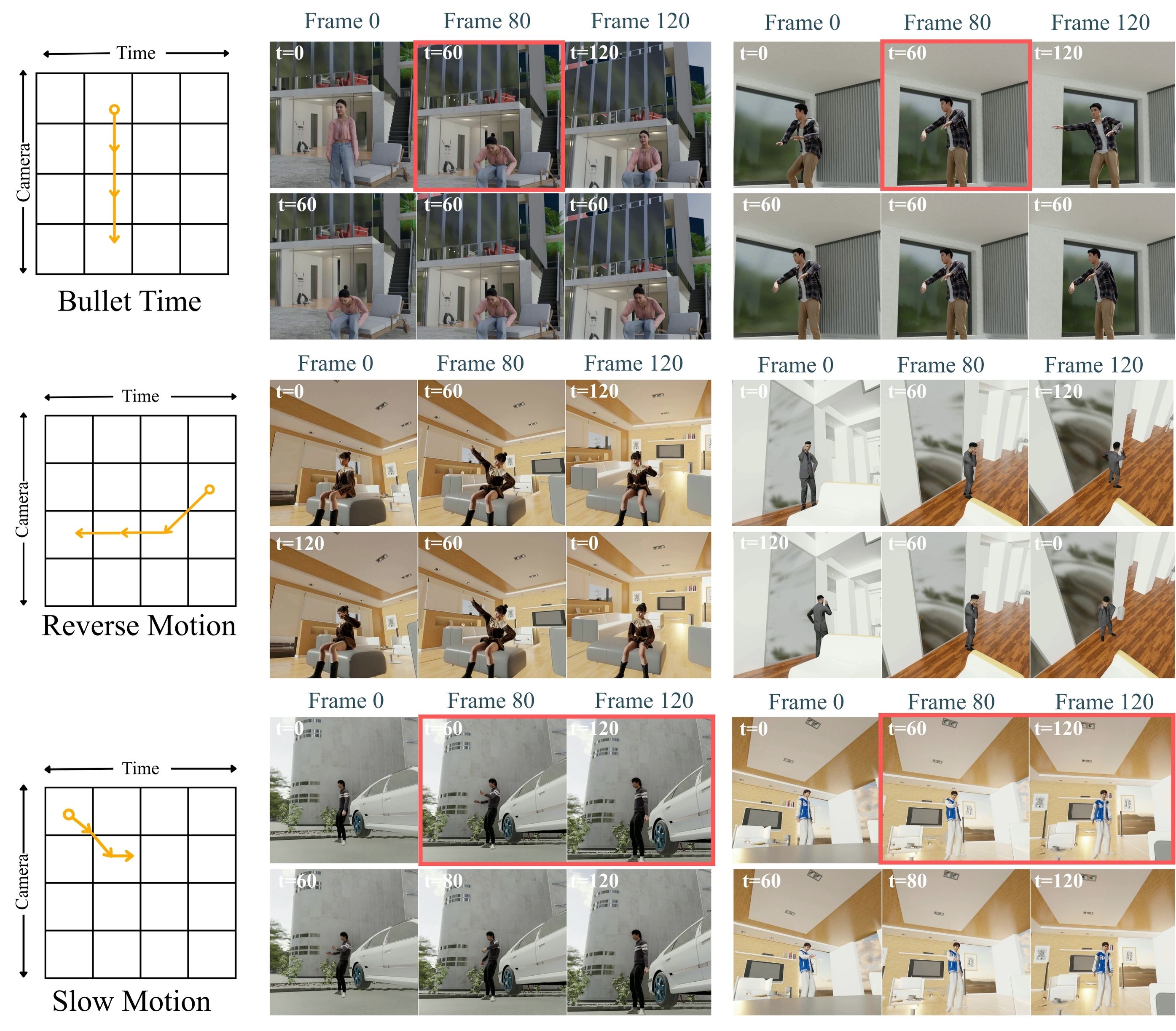}
    
    \caption{
    \textbf{Sampling from \nameDataset.}
By sampling from the \nameDataset dataset, we can extract frames corresponding to arbitrary combinations of camera viewpoints and temporal positions, forming source-target pairs with rich camera and temporal control signals.
    }
    \label{fig:sampling}
\end{figure}

For temporal-control training, we could sample any time variants from these sequences, including slow motion, reverse playback, bullet-time around arbitrary frames, and non-monotonic time patterns such as forward–backward oscillation. These augmented temporal signals are illustrated in \cref{fig:sampling}.

\begin{figure*}[ht]
    \centering
    \vspace{-3mm}
    \includegraphics[width=\linewidth]{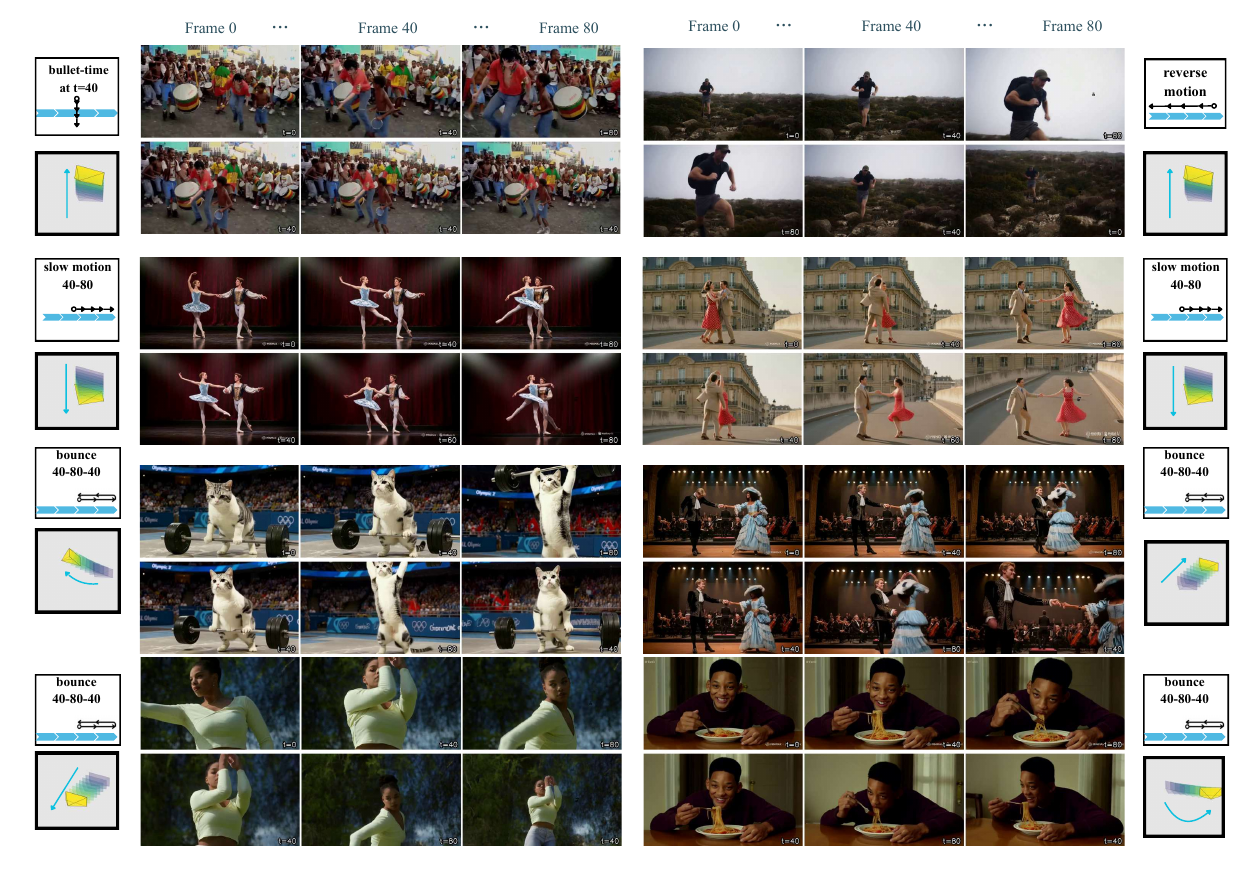}
    \caption{
    \textbf{More Qualitative results.}
Our model provides fully disentangled control over camera motion and temporal dynamics. Each row illustrates a different combination of temporal control inputs (top-left icon) and camera trajectories. \nameMethod consistently produces coherent videos across a wide range of controls, including normal and reverse playback, bullet-time, slow motion, replay motion, and complex camera paths such as panning, tilting, zooming, and vertical motion.
    }
    \label{fig:more_quali}
\end{figure*}

\section{Additional Ablation Studies}

\subsection{Temporal Warping Augmentation}
\begin{figure*}[ht]
    \centering
    \vspace{-3mm}
    \includegraphics[width=\linewidth]{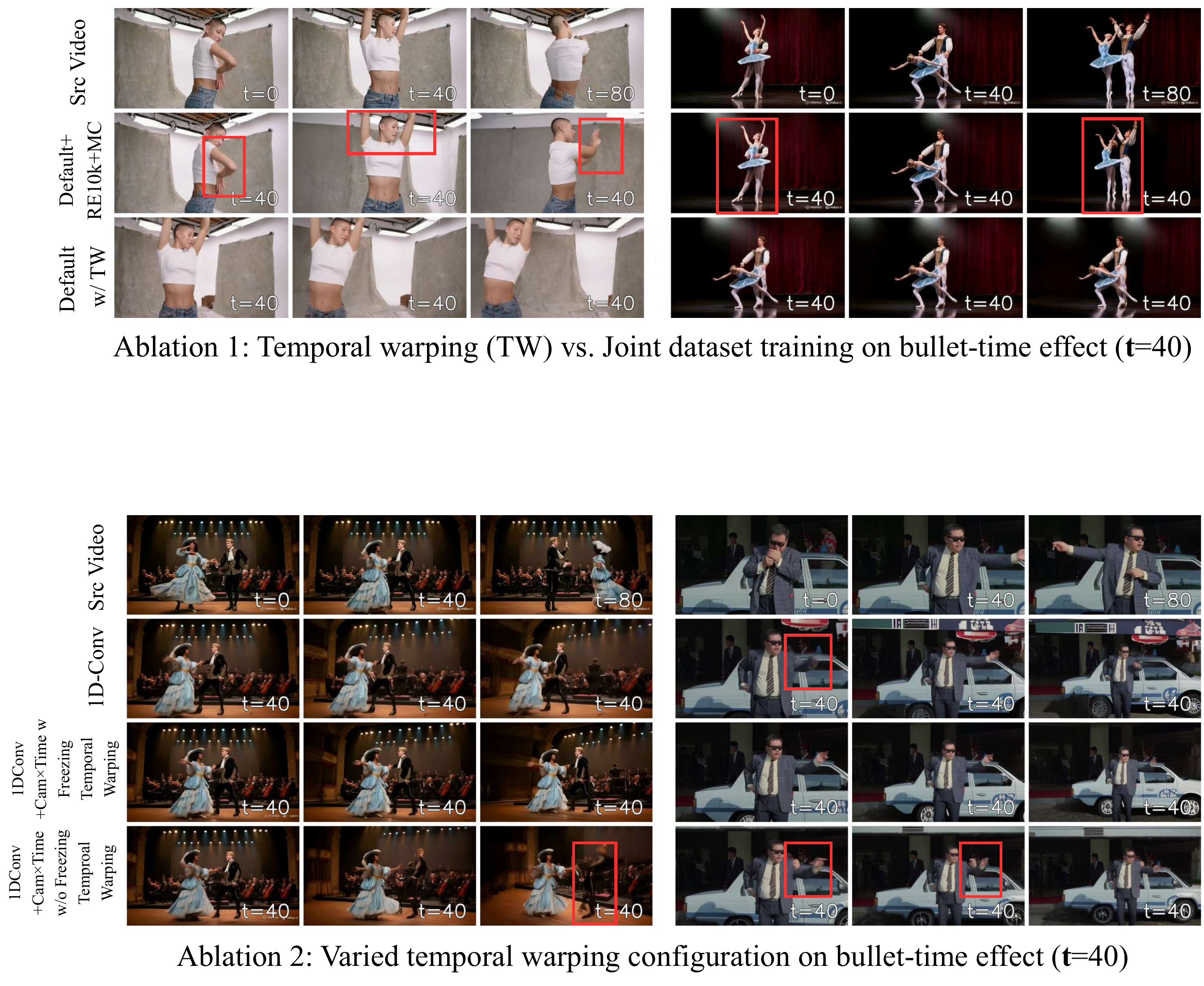}
    
    \caption{
    \textbf{Ablation study.}
    (Top) Using \cite{Bai2025,bai2024syncammaster} as default datasets, we compare the influence of adding static-scene datasets \cite{zhou2018stereo,li2019mannequin} vs.~just doing temporal warping (TW) augmentation (Sec.~3.2.2 in the main paper). Temporal warping definitely provide more variety of time control signals, allowing models to learn better camera-time disentanglement. 
    (Bottom) We further compare different configurations of warping, where we show freezing temporal warping (3\textsuperscript{rd} row) leads to better results than those trained without freezing temporal warping.
    }
    \label{fig:ablation_1_2}
\end{figure*}

Using \cite{Bai2025,bai2024syncammaster} as our default datasets, we compare training jointly with static-scene datasets \cite{zhou2018stereo,li2019mannequin} with applying only temporal warping (TW) augmentation on the default datasets (Sec.~3.2.2 in the main paper). Although static-scene datasets naturally support bullet-time effects, they do not provide enough diversity of temporal control configurations for models to reliably learn time locking on their own, as shown in \cref{fig:ablation_1_2} (top). Please refer to section ``Effective Temporal Warping'' in the website for more videos.

In \cref{fig:ablation_1_2} (bottom), we further show that freezing temporal warping (3rd row) produces better results than training without freezing it. Please refer to section ``Freeze Warping Ablations'' in the website for more videos. 

\subsection{Significance of \texttt{\nameDataset} Dataset}
\begin{figure*}[ht]
    \centering
    \vspace{-3mm}
    \includegraphics[width=\linewidth]{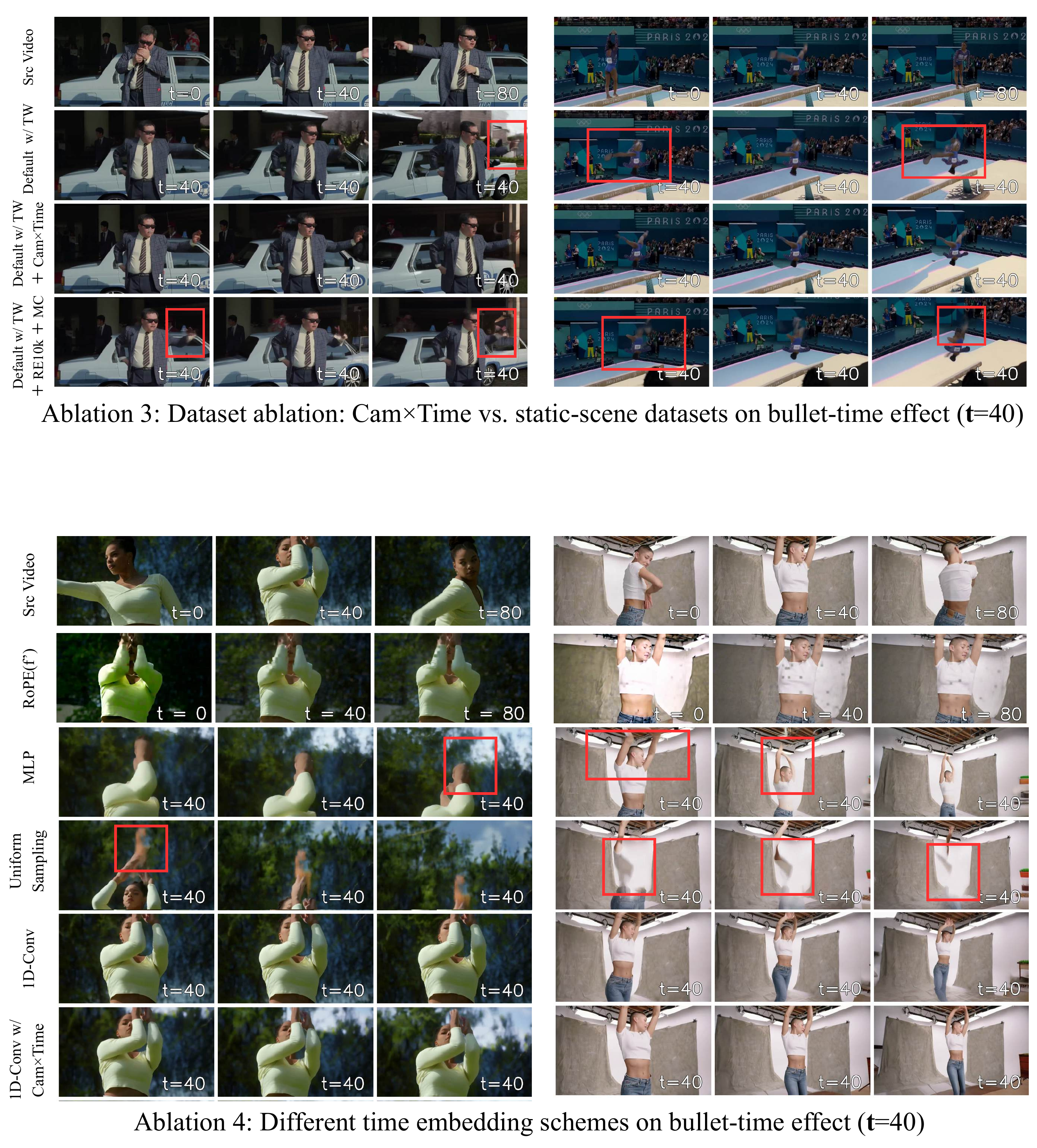}
    
    \caption{
    \textbf{Ablation study.}
    (Top) We verify the efficacy of the proposed \nameDataset dataset. Considering \cite{Bai2025,bai2024syncammaster} as default datasets, we compare the impact of different datasets on the generated videos. One can clearly see artifacts in baselines without any extra data or augmented with static-scene data, whereas training additionally with \nameDataset leads to no artifacts, confirming the usefulness of our dataset.
    (Bottom) We compare several time-embedding strategies. The MLP fails to lock the temporal state (red boxes), while RoPE($f'$) correctly freezes the scene dynamics at $\animatime$=40 but unintentionally locks the camera motion too. Conditioning on the latent frame $f'$ (with uniform sampling) introduces noticeable artifacts. In contrast, the proposed 1D-Conv embedding allows \nameMethod to both freeze the scene dynamics at $\animatime$=40 and produce intended camera motion. Incorporating \nameDataset during training further improves performance.
    }
    \label{fig:ablation_3_4}
\end{figure*}

Besides the quantitative results in the main paper (Table 5), in \cref{fig:ablation_3_4} (top), we provide visual comparisons demonstrating the effectiveness of the proposed \texttt{\nameDataset} dataset. 
Clear artifacts appear in baselines trained without additional data or with only static-scene augmentation (highlighted in red boxes), whereas incorporating \texttt{\nameDataset} removes these artifacts, demonstrating its significance. Please refer to section ``Dataset Ablations'' in the website for more videos.

\subsection{Time Embedding Ablation}

As promised in Sec.~3.2.1 in the main paper, we compare several time-embedding strategies. RoPE($f'$) can freeze the scene dynamics at $\animatime$=40, but it also undesirably locks the camera motion. 
Using MLP, by contrast, fails to lock the temporal state at all (red boxes).
Conditioning on the latent frame $f'$ (with uniform sampling) introduces noticeable artifacts. 
In comparison, the proposed 1D-Conv embedding enables \nameMethod to preserve the intended scene dynamics while still generating accurate camera motion. Adding \texttt{\nameDataset} to training further enhances the results. Please refer to section ``Time-Embedding Method Ablation'' in the website for more examples.

\section{Additional Qualitative Visualizations}
We show more qualitative results of \nameMethod in \cref{fig:more_quali}.
Our model provides fully disentangled control over camera motion and temporal dynamics. Each row presents a different pairing of temporal control inputs (top-left icon) and camera trajectories. \nameMethod reliably generates coherent videos under diverse conditions, including normal and reverse playback, bullet-time, slow motion, replay motion, and complex camera movements such as panning, tilting, zooming, and vertical translation. Please refer to section ``Video Demonstrations'' in the website for more examples. 

\clearpage

\end{document}